\documentclass[lettersize,journal]{IEEEtran}
\usepackage{amsmath,amsfonts}
\usepackage{algorithmic}
\usepackage{algorithm}
\usepackage{array}
\usepackage[caption=false,font=normalsize,labelfont=sf,textfont=sf]{subfig}
\usepackage{textcomp}
\usepackage{stfloats}
\usepackage{caption}
\usepackage{url}
\usepackage{verbatim}
\usepackage{graphicx}
\usepackage{cite}
\usepackage{multirow}
\usepackage{multicol}
\usepackage{colortbl}
\usepackage{xcolor}
\usepackage{enumitem}
\usepackage{hyperref}
\newcommand\blfootnote[1]{%
	\begingroup
	\renewcommand\thefootnote{}\footnotetext{#1}%
	\endgroup
}

\begin{document}

\title{Webly-Supervised Image Manipulation Localization via Category-Aware Auto-Annotation}

\author{Chenfan Qu, Yiwu Zhong, Huiguo He, Bin Li~\IEEEmembership{Senior Member,~IEEE}, Lianwen Jin~\IEEEmembership{Member,~IEEE}}

\markboth{Journal of \LaTeX\ Class Files,~Vol.~14, No.~8, August~2021}%
{Shell \MakeLowercase{\textit{et al.}}: A Sample Article Using IEEEtran.cls for IEEE Journals}

\twocolumn[{%
\renewcommand\twocolumn[1][]{#1}%
\maketitle 
\begin{center} 
\centering 
\vspace{-0.8cm}
\includegraphics[width=1.0\textwidth]{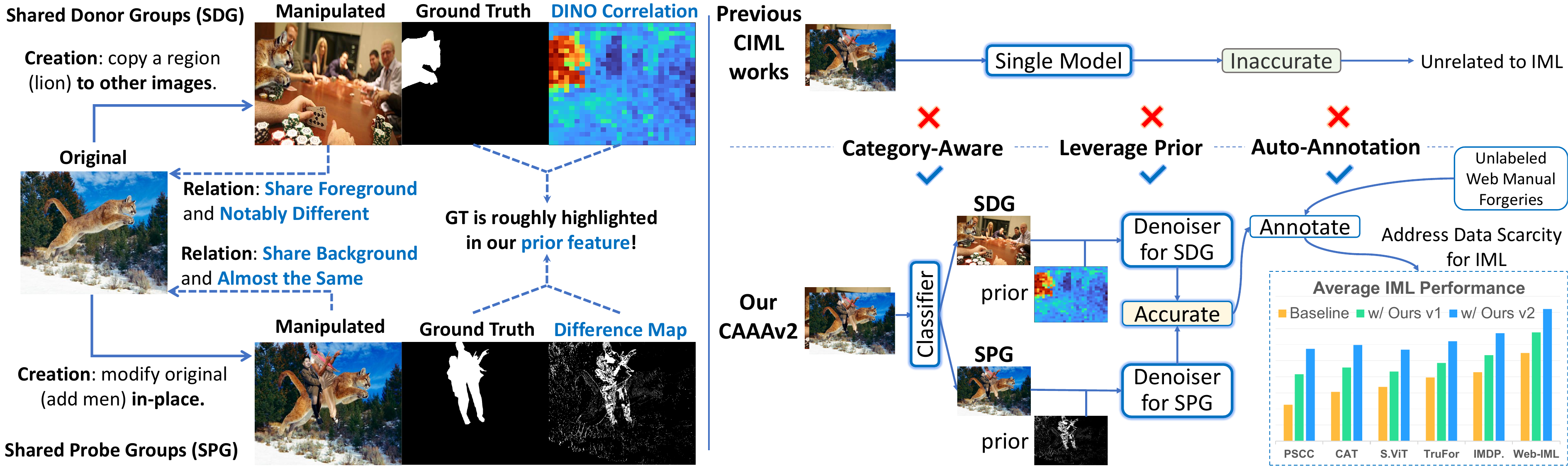} 
\captionof{figure}{\textbf{Left}: Definition and characteristics of the Shared Donor Group (SDG) and Shared Probe Group (SPG). \textbf{Right}: In contrast to previous works, our CAAAv2 adopts category-aware and prior-feature-denoising paradigm that notably reduces task difficulty and overfitting. We also propose using the CAAAv2 models to auto-annotate web-scale manually forged images. This approach significantly addresses the data scarcity issue for IML and notably improves the generalization of all IML models.} 
\label{fig: teaser}
\end{center}
}]

\blfootnote{This research is supported in part by
NSFC (Grant No.: 62476093) and GD-NSF (No.2021A1515011870). (\textit{Corresponding author: Lianwen Jin.})

Chenfan Qu, Huiguo He, and Lianwen Jin are with the School of Electronics and Information Engineering, South China University of Technology, Guangdong 510641, China.
(Email: 202221012612@mail.scut.edu.cn, hehuiguo@scut.edu.cn, eelwjin@scut.edu.cn)

Yiwu Zhong is with the School of Intelligence Science and Technology, Peking University, Beijing 100091, China. (Email: yiwu-zhong@outlook.com)

Bin Li is with the College of Electronics and Information Engineering, Shenzhen University, Shenzhen 518060, China. (Email: libin@szu.edu.cn)
}

\begin{abstract}
Images manipulated by image editing tools can mislead viewers and pose significant risks to social security. However, accurately localizing manipulated image regions remains challenging due to the severe scarcity of high-quality annotated data, which is laborious to create.
To address this, we propose a novel approach that mitigates data scarcity by leveraging readily available web data. We utilize a large collection of manually forged images from the web, as well as automatically generated annotations derived from a simpler auxiliary task, constrained image manipulation localization.
Specifically, we introduce \textbf{CAAAv2}, a novel auto-annotation framework that operates on a category-aware, prior-feature-denoising paradigm that notably reduces task complexity.
To further ensure annotation reliability, we propose \textbf{QES}, a novel metric that filters out low-quality annotations. Combining CAAAv2 and QES, we construct \textbf{MIMLv2}, a large-scale, diverse, and high-quality dataset containing 246,212 manually forged images with pixel-level mask annotations. 
This is over 120$\times$ larger than existing handcrafted datasets like IMD20.
Additionally, we introduce \textbf{Object Jitter}, a technique that further enhances model training by generating high-quality manipulation artifacts. 
Building on these advances, we develop \textbf{Web-IML}, a new model designed to effectively leverage web-scale supervision for the task of image manipulation localization.
Extensive experiments demonstrate that our approach substantially alleviates the data scarcity problem and significantly improves the performance of various models on multiple real-world forgery benchmarks. 
With the proposed web supervision, our Web-IML achieves a striking performance gain of 31\% and surpasses the previous state-of-the-art SparseViT by 21.6 average IoU points.
The dataset and code will be released at https://github.com/qcf-568/MIML.

\end{abstract}


\section{Introduction}
\IEEEPARstart{T}{he} misuse of manipulated images can lead to fraud and the spread of misinformation, posing serious threats to social media security~\cite{pami, pami6, pami7, pami8, pami9, pami10, pami11, pami+}. 
To this end, Image Manipulation Localization (IML) has garnered increasing attention in recent years~\cite{pami1, pami2, pami3, pami4, pami5, pami++, tifsiml1, tifsiml2}. 
A central challenge in IML lies in the scarcity of large-scale high-quality data~\cite{ncliml}, which is essential for preventing model overfitting. Creating such data is extremely labor-intensive and time-consuming~\cite{sparsevit}, requiring not only the crafting of convincing forgeries but also the precise pixel-level annotation of manipulated regions.
Although alternatives like synthetic data~\cite{catnet, defacto} or AI-generated forgeries~\cite{wang2025opensdi} have been explored, they often exhibit significant discrepancies from real-world manipulations, thereby limiting model generalization to practical scenarios~\cite{ncliml}.

Meanwhile, numerous unlabeled, manually forged images are publicly available over the Internet, often accompanied by their corresponding originals. 
To address the scarcity of annotated data, we propose to leverage these unlabeled forgeries by automatically generating pixel-level mask annotations using Constrained Image Manipulation Localization (CIML) models.
Our key insight is that CIML can reduce the complexity of manipulation localization by comparing the forged image with its original, resulting in more accurate and reliable annotations. Therefore, CIML offers a practical solution for automatically generating large-scale, high-quality training datasets for IML.

However, despite recent progress with CIML on simple synthetic images, existing methods are unreliable for complex, real-world forgeries due to two major limitations:

\textbf{First}, previous methods employ a suboptimal paradigm that relies on a single correlation-based model for all input types~\cite{ciml1, ciml2}. In practice, image pairs consisting of forged images and their original counterparts can be categorized into two distinct groups (see Fig.~\ref{fig: teaser}): (1) \textbf{Shared Donor Group (SDG)} denotes image pairs that share \underline{foreground}, with forged images generated by copying objects from the original image; (2) \textbf{Shared Probe Group (SPG)} denotes image pairs that share \underline{background}, where forged images are obtained by modifying the original image in-place. While correlation-based methods perform reasonably well on SDG data, they falter on SPG data because the shared region (the background) often lacks distinctive features. Forcing a single model to handle both groups leads to confusion and degraded performance.

\textbf{Second}, the CIML task itself is challenging and also suffers from a scarcity of high-quality training data. Previous works provide the model with only the image pair, relying entirely on the network to learn the complex correspondence between the images in each pair. Without clear and explicit guidance, models struggle to learn this correspondence effectively and are prone to overfitting on limited training data.

To overcome these challenges, we propose a novel paradigm: \textbf{Category-Aware Auto-Annotation v2 (CAAAv2)}. It achieves superior generalization via two key innovations:

(1) \textbf{Category awareness}. We treat image pairs in SDG and SPG separately through a classifier at start. This separation avoids the confusion of joint training and allows us to deploy distinct, optimized pipelines that leverage the unique correspondence of each category. This classifier can be effectively trained on unlabeled images. 

(2) \textbf{Task transformation}. We reframe the challenging CIML problem as a far simpler denoising task. This is achieved by introducing novel prior features that provide strong initial guidance for localization. Crucially, these prior features are extracted via zero-shot frozen functions that are resistant to overfitting. For SDG pairs, where the forged regions are typically shared objects, we extract features from both images using a frozen DINOv2 encoder and compute their correlation map. The resulting map, named DINO correlation, serves as a prior that roughly highlights the manipulated regions (Fig.~\ref{fig: teaser}). The model's task is reduced to simply denoising this prior to produce a precise mask. For SPG pairs, where the forged regions are the difference between the images, we obtain the prior map by subtracting the forged image from its original. Through task transformation, we significantly reduce task difficulty and bridge the gap between synthetic and real-world data, leading to substantially better generalization.

Subsequently, we collect a large amount of manually forged images from the Internet and then annotate their forged regions with the proposed CAAAv2. This approach considerably alleviates the data scarcity in image manipulation localization. To ensure the reliability of the annotations, we propose Quality Evaluation Score (\textbf{QES}), a novel metric that automatically assesses and filters out inadequate annotations.

While this pipeline yields a high-quality dataset, we observed a relative lack of sophisticated copy-move forgeries that feature only subtle edge artifacts. Direct copy-move synthesis is not effective since it brings obvious semantic artifacts. To address this, we introduce \textbf{Object Jitter}, a data synthesis method that simulates these challenging cases by applying minor perturbations (e.g., enlargement) to random objects on authentic images, effectively supplementing our dataset.

In our CVPR 2024 paper, we propose to address the data scarcity of IML by auto-annotating images from the web with a category-aware approach CAAA. However, CAAA did not incorporate a prior-feature-denoising paradigm for its SDG branch, leading to overfitting on limited training data. In this paper, we propose \textbf{CAAAv2, an improved version of our CAAA} that resolves this issue by leveraging frozen DINO-based prior features. The enhanced accuracy of CAAAv2 ensures that a greater number of samples are retained after filtering. Consequently, the collected dataset \textbf{MIMLv2 is both larger and more diverse than our conference version MIMLv1}~\cite{mimlv1}, better supporting the development of IML models (Figure~\ref{fig: teaser}). Moreover, different from our conference paper that only uses MIMLv1 dataset to improve IML models, \textbf{our Object Jitter approach further effectively supplements MIMLv2 with high-quality hard samples}. 

To better leverage our high-quality web-scale supervision for IML, we further propose a new model named Web-IML, which comprises a Multi-Scale Perception module and a Self-Rectification module. The former enables the model to better integrate information from multiple perspectives; the latter allows the model to learn to identify and correct its mistakes. Our Web-IML significantly outperforms previous methods on various widely used benchmarks, \textbf{being smaller yet more effective than the APSC-Net in our conference version}.

In summary, our main contributions are as follows:
\begin{itemize}
\item We propose to address the scarcity of IML training data by leveraging the less challenging CIML task to auto-annotate forged images collected from the web.
\item We introduce \textbf{CAAAv2}, a novel CIML paradigm that significantly enhances generalization. This is achieved through a category-aware design, which treats distinct image groups separately, and a prior-denoising approach, which reframes CIML as a simpler denoising task.
\item With CAAAv2 and a novel metric QES, we automatically annotate manually forged web images, constructing a large-scale, diverse, high-quality dataset \textbf{MIMLv2}. This dataset considerably alleviates the data scarcity in IML.
\item We introduce \textbf{Object Jitter}, which generates high-quality, challenging-to-localize artifacts, effectively supplementing our dataset.
To better leverage the web-scale supervision, we design a novel model \textbf{Web-IML}. Extensive experiments have validated its effectiveness.
\end{itemize}

 \begin{figure*}[t]
 	\centering
 	\setlength{\abovecaptionskip}{-0.1cm}
 	\includegraphics[width=1.0\linewidth]{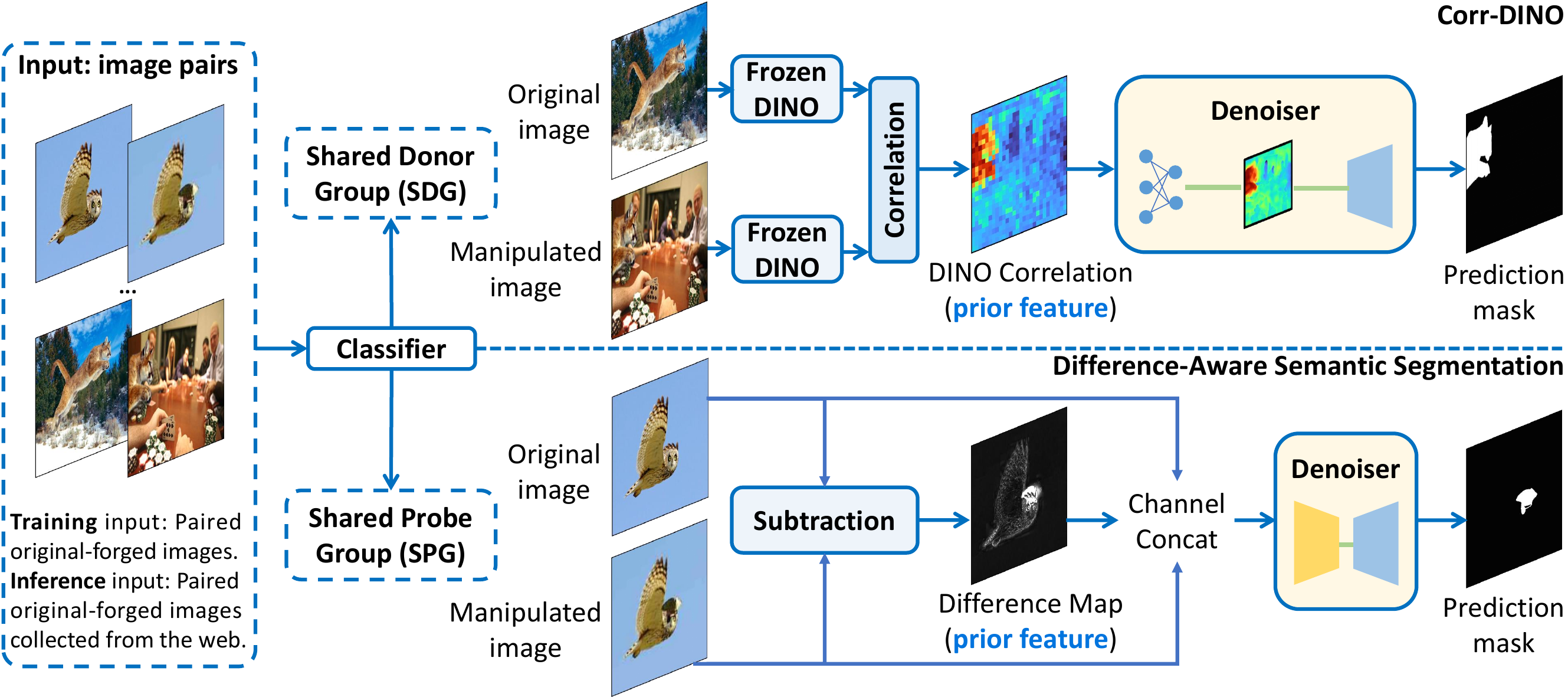}
 	\centering
 	\caption{The overall pipeline of our proposed CAAAv2 paradigm. The input original-forged image pairs are first categorized as either SDG or SPG with a classifier. Then, we use Corr-DINO and Difference-Aware Semantic Segmentation to process SDG and SPG pairs respectively. Both models first extract a prior feature that roughly highlights forgeries and then denoise it.
 	} 
\label{fig:Fig3_CAAA}
 \end{figure*}

\section{Related Works}
\subsection{Image Manipulation Localization (IML)}
Image Manipulation Localization aims to identify forged regions within images~\cite{yu2024diffforensics, napl, imla1, imla2, fakeshield, imla3, liuyq, imla4, imla5}. The scale of existing handcrafted datasets is limited due to the high cost of manual collection and annotation, inevitably leading to overfitting~\cite{sparsevit, ncliml}. To mitigate this issue, some studies have incorporated handcrafted features. For instance, RGB-N~\cite{zhou2018learning} applied SRM~\cite{zhou2018learning} filters to suppress semantic bias. MVSS-Net~\cite{dong2022mvss} and MGQFormer~\cite{zeng2024mgqformer} further exploited learnable noise domain modeling. CAT-Net~\cite{catnet}, ObjectFormer~\cite{wang2022objectformer} and HiFi-Net~\cite{hifi} leveraged frequency features to localize manipulations. However, the handcrafted features are noisy and unstable, hindering further improvements in model performance. Recently, pure vision models have made notable progress~\cite{imdl3, pami}. NCL~\cite{ncliml} employed contrastive learning to improve generalization. SparseViT~\cite{sparsevit} utilized dilated attention to reduce semantic bias. PIM-Net~\cite{pami} explicitly modeled the super-pixel inconsistency to alleviate biased learning. Despite these advances, overfitting remains a significant challenge due to the fundamental issue: the scarcity of high-quality training data.

To address this fundamental issue, we propose to construct a large-scale, high-quality dataset by automatically annotating Internet images with a less challenging task, CIML.


\subsection{Constrained Image Manipulation Localization (CIML)} 
Constrained Image Manipulation Localization aims to identify forged regions in an image by leveraging its original counterpart~\cite{dmvn}. Previous works relied on correlation matching and handled SDG and SPG image pairs uniformly. For instance, DMVN~\cite{dmvn} computed correlation maps to localize similar objects across image pairs. DMAC~\cite{ciml1} incorporated dilated convolution to enhance spatial information. AttentionDM~\cite{ciml2} employed an attention mechanism to improve performance. MSTAF~\cite{cimlmm} performed correlation in both the encoder and decoder to extract richer features. 
While these methods achieved notable progress on less challenging datasets (e.g., synthetic COCO~\cite{dmvn}), their performance is limited in real-world forgeries, which often involve high-resolution images, significant variations, and complex details. 

In contrast, our approach reduces task difficulty and addresses overfitting by (1) explicitly harnessing the distinct characteristics of SPG and SDG pairs; and (2) transforming the CIML problem into a denoising task through prior features extracted from frozen functions.

\section{Category-Aware Auto-Annotation v2}\label{sec. 3}

\subsection{Motivation and Framework Overview}
For CIML, previous works overlooked the critical discrepancy and characteristics between SPG and SDG images, treating them using a single fully-tunable correlation model. We argue that this paradigm is suboptimal for two reasons:

\smallskip

1. \textbf{Conflicting Correlation Signals}. The nature of the "shared region" differs drastically between the two categories. In SDG pairs, the shared region is salient foreground objects with distinctive features (e.g., the lion in SDG of Fig.~\ref{fig:Fig3_CAAA}), providing a strong signal for correlation matching. Conversely, in SPG pairs, the shared region is the background, which lacks the unique features required for reliable matching (e.g., the sky in SPG of Fig.~\ref{fig:Fig3_CAAA}). Forcing a single correlation model to simultaneously handle both groups creates a conflicting objective that degrades performance.

\smallskip

2. \textbf{Lack of Explicit Guidance}. Learning to comprehend and fully leverage the correspondence between original and manipulated images is non-trivial. With scarce training data, CIML models tend to find shortcuts to minimize training loss, often overfitting to either forgery-agnostic semantics in small-scale handcrafted datasets or monotonous patterns in synthetic data. Without explicit guidance on how to utilize the correspondence within an image pair, models often fail to generalize to diverse, real-world manipulations.

\smallskip

To overcome these challenges, we propose \textbf{Category-Aware Auto-Annotation v2 (CAAAv2)}, a novel paradigm for CIML. Our \textbf{core idea} is to process SDG and SPG images independently, while explicitly integrating image correspondence as a guiding prior feature. Specifically, as shown in Fig.~\ref{fig:Fig3_CAAA}, CAAAv2 introduces a classifier to categorize the input as either SPG or SDG, and then applies a category-specific prior to guide localization. For SDG pairs, we introduce DINO Correlation, a prior that captures shared foreground objects by correlating dense features from a frozen DINOv2 encoder. For SPG pairs, a simple Difference Map effectively identifies manipulated regions by subtracting the two images.

These priors embed domain-specific knowledge about forgery patterns into strong, zero-shot signals. By generating these priors with frozen, non-trainable functions, we reframe the complex task of open-ended forgery localization into a simpler, more constrained denoising problem. Here, the model's objective is to refine the coarse prior into a precise mask, using the input images as context. This paradigm shift significantly reduces task difficulty and minimizes the generalization gap between synthetic and real-world forgeries, leading to substantial performance improvements.

Importantly, the trained models are further used to auto-annotate large-scale, unlabeled manual forgeries from the web, thereby addressing the data scarcity issue in IML.

Whereas CAAA overfits on SDG data, CAAAv2 provides explicit guidance by introducing DINO Correlation as a strong prior. This prior is generated using a frozen DINOv2, whose global receptive field naturally aligns multi-scale semantics, yielding substantial gains in real-world generalization.

\subsection{Self-Supervised Classification}
\label{subsec:classifier}
Training a classifier to distinguish between SDG and SPG is a non-trivial task due to the lack of labeled data. To address this challenge, we propose training a classifier via self-supervised learning on unlabeled images. Given an image, we apply random augmentations (e.g.,~color jitter, image compression) and in-place manipulations (copy-paste, splicing, or removal). The manipulated image and the original image form an SPG pair. To construct an SDG pair, we copy 1-3 random regions from the original image, resize them, and paste them into another image. With the obtained image pairs, we can effectively train our classifier. 
Our classifier actually only needs to discern whether the main area of the input image pair is identical or not, without considering which image or region is fake. For example, in Fig.~\ref{fig:Fig8_MIML}, the two images in any SPG pair are nearly identical and spatially aligned, whereas images in any SDG pair are distinctly different. These characteristics are universally applicable and strong enough. Therefore, this classification task is quite simple, enabling accurate separation of images into the two groups.

\subsection{Corr-DINO}
\label{subsec:DINO}
\begin{figure*}[t]
 	\centering
 	\includegraphics[width=1.0\linewidth]{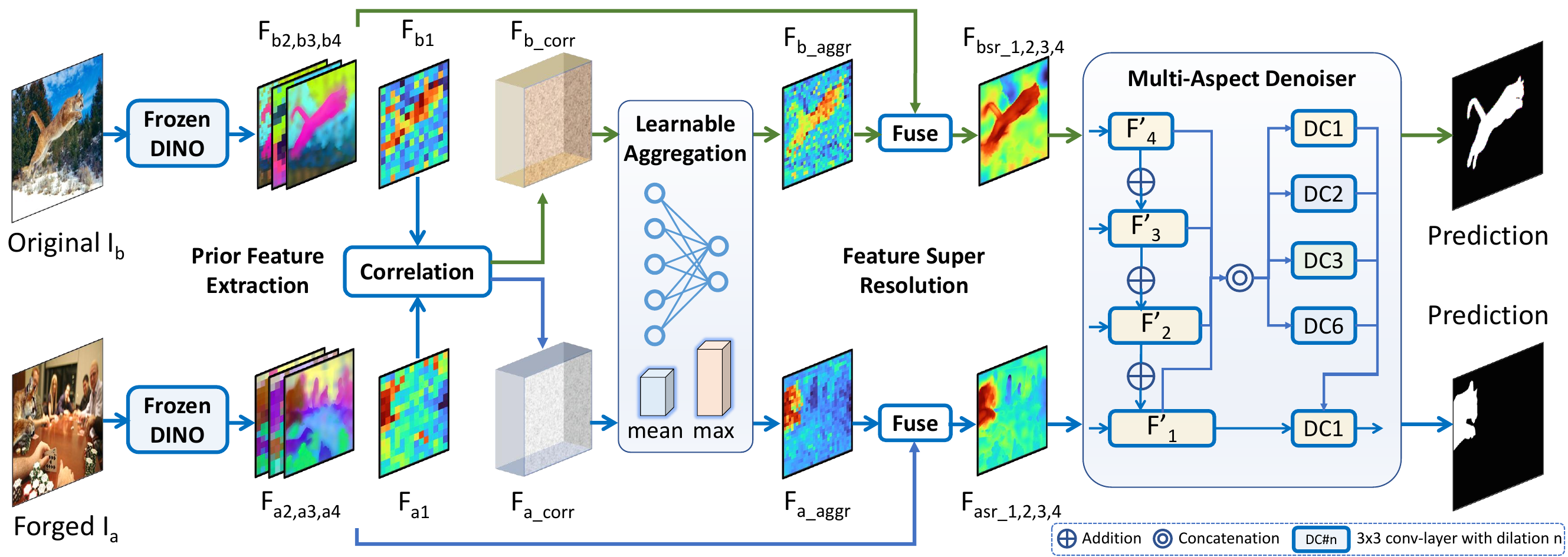}
 	\centering
 	\setlength{\abovecaptionskip}{-0.4cm} 
 	\caption{
    The overall pipeline of our proposed Corr-DINO. First, we employ a frozen DINO backbone to extract image features. The features obtained from the final DINO layer are then used for correlation calculation. Next, a learnable aggregation module reduces the channel dimension of the correlation features, which are further refined using a Feature Super Resolution module. Finally, the refined features are denoised with a Multi-Aspect Denoiser, resulting in the final mask prediction. Green lines indicate operations used only during training with synthetic data, due to the lack of mask labels for the real images.} \label{fig:Fig6_CorrDINO}
 \end{figure*}

For Shared Donor Group (SDG) pairs, the manipulated regions are mostly the objects copied from the original image. The localization task therefore reduces to identifying these shared foreground objects. To exploit this correspondence, we introduce a novel prior feature, termed DINO Correlation. This prior is generated by extracting dense features from both images using a frozen DINOv2 encoder and then computing their cross-correlation. The resulting map provides a strong yet coarse signal that highlights the shared objects, which often are the forged regions. The task is thus simplified into refining this coarse prior to a precise segmentation mask. Based on this, we propose the Corr-DINO model. As illustrated in Figure~\ref{fig:Fig6_CorrDINO}, it consists of four components:

\vspace{+0.1cm}
\smallskip

\noindent 1. \textbf{Prior Feature Extractor} for prior feature extraction. We extract object features from both the tampered image $I_{a}$ and its original image $I_{b}$ with a frozen DINOv2~\cite{oquab2023dinov2} backbone, which has learned generalized representations through self-supervised learning on vast-scale web images. The features from the last DINO layer, denoted as $F_{a1}$ and $F_{b1}$, contain rich semantic information and are used to compute correlation features $F_{a\_corr}$ and $F_{b\_corr}$. $F_{a\_corr} = [Corr(F_{a1}, F_{b1}), Corr(F_{a1}, F_{a1})]$, $F_{b\_corr} = [Corr(F_{b1}, F_{a1}), Corr(F_{b1}, F_{b1})]$. The same correlation function $Corr$ as most previous works~\cite{ciml2, cimlmm, mimlv1} is adopted.

\vspace{+0.1cm}
\smallskip

\noindent 2. \textbf{Learnable Aggregation} module for dimension reduction. Previous works simply select the top-K activated channels with the highest average values in $F_{a\_corr}$ and $F_{b\_corr}$ to reduce dimension. 
This approach works for trainable encoders but is suboptimal for our frozen encoder, whose correlated object features are relatively sparse. 
Selecting only the top-K channels may discard important features and compromise the integrity of $F_{a\_corr}$ and $F_{b\_corr}$.
To overcome this, we propose a learnable aggregation module that uses trainable parameters to reduce the channel count to K. We also include statistics of average and maximum values across the channel dimension. $F_{a\_aggr}=[C_{1\times1}(R(C_{1\times1}(F_{a\_corr}))), Avg(F_{a\_corr}),Max(F_{a\_corr})]$. Here, $C_{1\times1}$ is $1\times1$ conv-layer, $R$ is ReLU. $Avg$ and $Max$ are average and maximum pooling respectively. A similar process is used for $F_{b\_corr}$ to generate $F_{b\_aggr}$.

\vspace{+0.1cm}
\smallskip

\noindent 3. \textbf{Feature Super Resolution} module for image detail reconstruction. The features from the last DINOv2 layer are rich in semantics but lack fine image details~\cite{oquab2023dinov2}, making $F_{a\_aggr}$ and $F_{b\_aggr}$ too coarse for fine-grained mask predictions. To reconstruct image details, we leverage features from the last 2-4 layers of DINO, which are known to be more sensitive to edges and fine details~\cite{oquab2023dinov2}. We concatenate the features from the last four encoder layers $F_{a}=Concat(F_{a1,2,3,4})$ and interpolate $F_{a\_aggr}$ to different scales ($1\times,2\times,4\times,8\times$) to obtain $F_{aggr\_a1,2,3,4}$. We then fuse them with $1\times1$ conv-layers. A similar process is applied to $F_{b\_aggr}$.

\vspace{+0.1cm}
\smallskip

\noindent 4. \textbf{Multi-Aspect Denoiser}\label{MAD} for feature denoising. Since the DINOv2 encoder is frozen, its output features inevitably contain significant task-irrelevant noise.
To achieve precise mask prediction, we propose denoising the correlation features by integrating multi-perspective information. This module fuses multi-scale features $F_{asr\_1,2,3,4}$ in a top-down manner. The fused features are further concatenated and denoised using a set of dilated conv-layers. The output is concatenated to $F'1$, which is used to yield the mask prediction. More details are provided in the Appendix.



 \begin{figure}[t]
 	\centering
 	\includegraphics[width=1.0\linewidth]{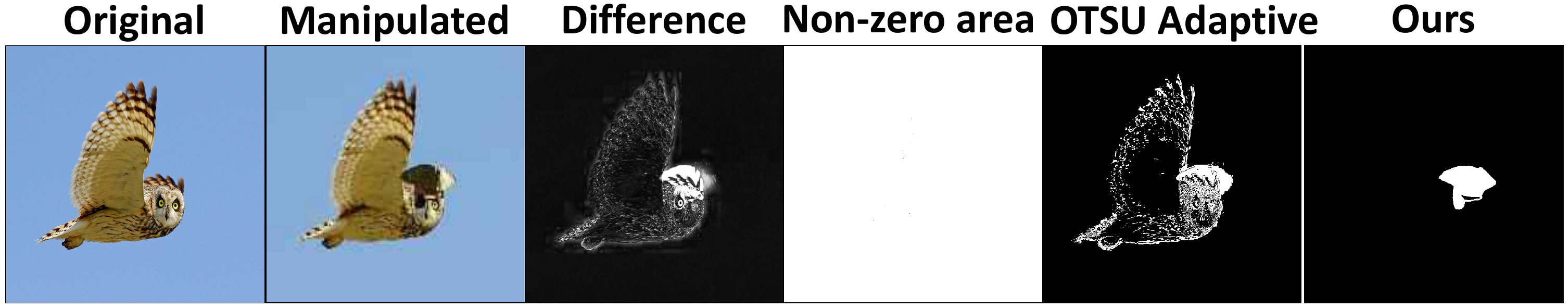}
 	\centering
 	\setlength{\abovecaptionskip}{-0.5cm} 
 	\caption{Manipulated images often degrade during transmission. As a result, the absolute difference between a tampered image and its original does not accurately indicate the forged region. Our method effectively denoises the difference map by leveraging semantic information.
 	} \label{fig:Fig4_Bear}
 \end{figure}

\subsection{Difference-Aware Semantic Segmentation (DASS)}
\label{subsec:DASS}
SPG pairs are characterized by in-place manipulation, where the manipulated image is generated by directly altering its original. This process inherently ensures precise spatial alignment and a shared background, meaning that the forged regions are always highlighted by the image difference. Leveraging this strong correspondence, we use the difference map as a prior feature to guide model learning. Given a manipulated image $I_m$ and its original $I_o$, the absolute difference map $D_{mo}=abs(I_m-I_o)$ can roughly highlight forged regions. 

However, transmission noise and image degradation often contaminate the difference map, making it too noisy for precise localization. As illustrated in Fig.~\ref{fig:Fig4_Bear}, nearly every area exhibits nonzero values due to noise, and even adaptive binarization~\cite{otsu} fails to denoise.
To address this issue, we propose \textbf{denoising the difference map by incorporating semantic information from the images}. Specifically, we concatenate $I_m$, $I_o$, and $D_{mo}$ along the channel dimension, and feed them into a semantic segmentation model. This model integrates a VAN~\cite{van} encoder for extracting rich semantic features and employs the same Multi-Aspect Denoiser (\ref{MAD}) as in our Corr-DINO for denoising the difference map.

\section{MIMLv2 Dataset}
In this section, we introduce \textbf{MIMLv2}, a large-scale, diverse, and high-quality dataset. \textbf{The key idea is to leverage CIML models trained on existing datasets to auto-annotate manual forgeries collected from the web, thereby addressing the data scarcity in IML.} To ensure high annotation quality, we propose a novel metric that filters out inadequate annotations without requiring ground-truth for evaluation.

\smallskip

Our MIMLv2 is annotated using our CAAAv2, which is significantly more accurate than its conference version, CAAA~\cite{mimlv1}. As a result, more samples are retained as high quality after filtering, as shown in Table~\ref{tab: datas}. Compared to MIMLv1~\cite{mimlv1}, MIMLv2 offers greater scale and diversity, making it a more effective resource for training IML models.

\begin{figure}[t]
 	\centering
 	\includegraphics[width=1.0\linewidth]{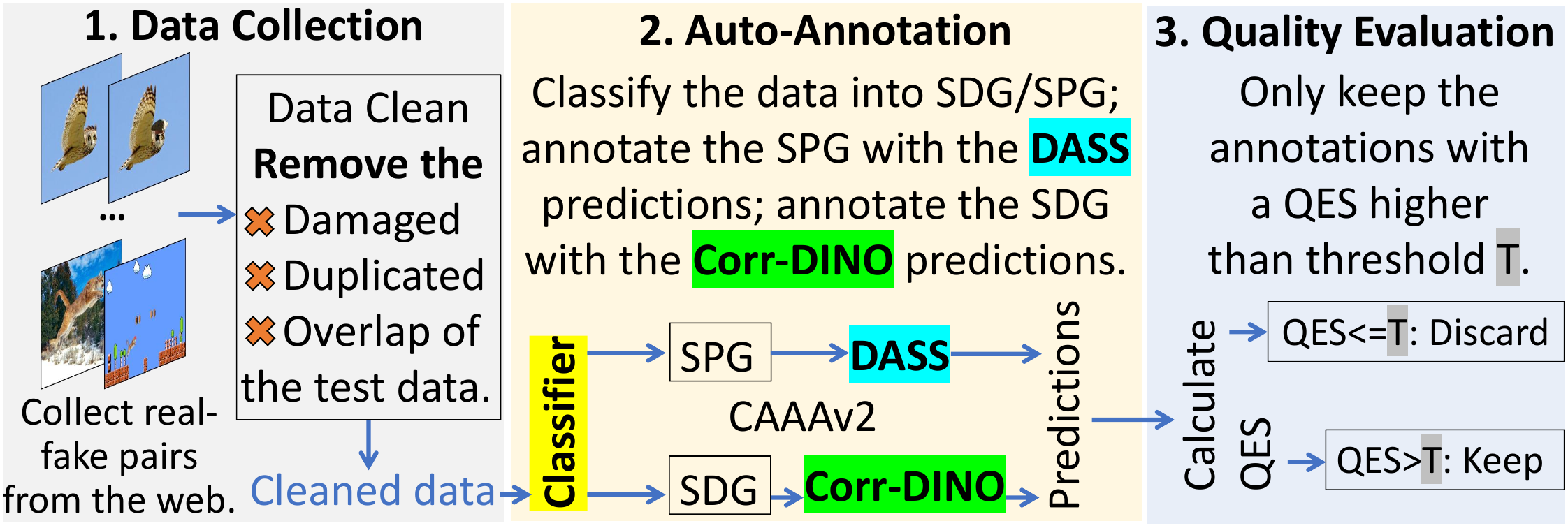}
 	\centering
 	\setlength{\abovecaptionskip}{-0.4cm}
 	\caption{The construction pipeline of our MIMLv2 dataset. 
 	} \label{fig:Fig7_dataconstruct}
 \end{figure}

\subsection{Dataset Construction}
As illustrated in Fig.~\ref{fig:Fig7_dataconstruct}, we construct MIMLv2 through the following steps:

\smallskip

\noindent\textbf{1. Image Collection.}~We collect forged images and their corresponding originals from imgur.com. On this website, the images are manually manipulated and thus have high-quality, diverse forged regions. Additionally, we exclude damaged, duplicated images and those that already appear in the evaluation datasets~\cite{imd20 ,casia, coverage, nist16, guillaro2023trufor, cimd, misd} with MD5~\cite{md5} and pHash~\cite{phash}.

\smallskip

\noindent\textbf{2. Auto-Annotation.}~We utilize the CAAAv2 models, as proposed in Section~\ref{sec. 3}, to obtain mask annotations for the collected images. The models are effectively trained on synthetic COCO~\cite{dmvn}, CASIAv2~\cite{casia} and IMD20~\cite{imd20} datasets.

\smallskip

\noindent\textbf{3. Quality Evaluation.} 
After auto-annotation, SPG annotations are already high-quality, whereas a subset of SDG annotations remains unsatisfactory. We empirically found that inaccurate predictions mostly have smooth edges and low confidence. Inspired by this, we propose \textbf{Quality Evaluation Score} (\textbf{QES}), a novel metric that assesses both confidence and edge sharpness to evaluate annotation quality and filter out inadequate ones.
Specifically, given a prediction mask with shape (H, W) and normalized probability, we compute the QES as follows: QES=$\frac{\sum_{i,j}^{H,W}p_{i,j}>(1-T_h)}{\sum_{i,j}^{H,W}p_{i,j}>T_l}$, where $\sum_{i,j}^{H,W}p_{i,j}>(1-T_h)$ denotes the area of prediction with a high confidence greater than $(1-T_h)$, and $\sum_{i,j}^{H,W}p_{i,j}>T_l$ denotes the total predicted potentially manipulated area. We empirically set $T_h$ and $T_l$ to $\frac{1}{16}$ and only retain samples with QES$>$0.5. In this way, QES can effectively filter unreliable mask annotations.

 \begin{figure*}[t]
 	\centering
 	\includegraphics[width=1.0\linewidth]{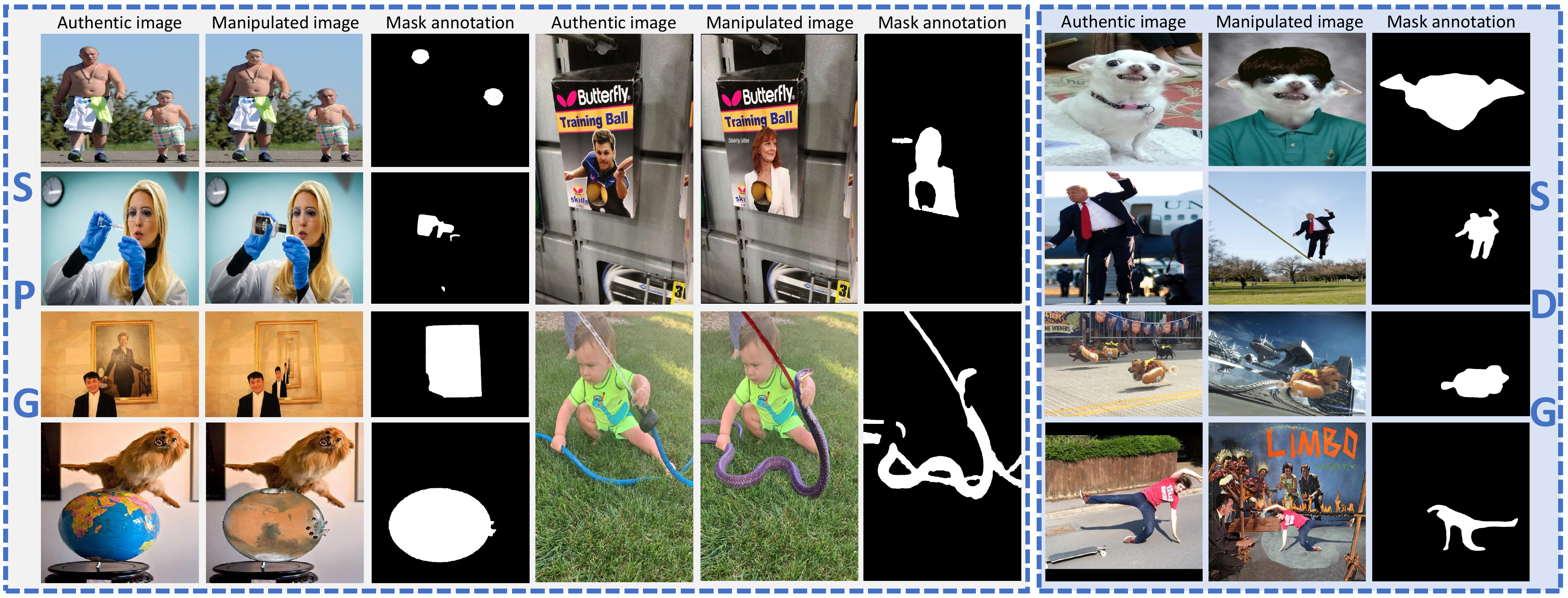}
 	\centering
 	\setlength{\abovecaptionskip}{-0.5cm}
 	\caption{Some images and their corresponding mask annotations from the proposed MIMLv2 dataset.
 	} \label{fig:Fig8_MIML}
 \vspace{-0.1cm}
 \end{figure*}

\subsection{Dataset Highlights}
We present a few samples of our MIMLv2 dataset in Fig.~\ref{fig:Fig8_MIML}. The forged regions are precisely annotated with high-quality masks, demonstrating the effectiveness of our auto-annotation approach. The main highlights of our dataset include:

\begin{itemize}[leftmargin=*]
\item \textbf{High quality.}~The proposed dataset contains high-quality image manipulations crafted by humans. Such data is crucial to training models that can detect real-world forgeries rather than overfit to synthetic data.

\item \textbf{Large Scale.} The proposed dataset is substantially larger than existing handcrafted IML datasets. As shown in Table~\ref{tab: datas}, it comprises 246,212 manually forged images, which is about 120 times more than IMD20~\cite{imd20}.

\item \textbf{Broad Diversity.} 
MIMLv2 comprises images of various sizes, styles, and manipulation types (e.g., copy-move, splicing, removal, AIGC), created by more than 10k individuals and various software. This diversity significantly enhances the generalization of IML models.

\item \textbf{Updated Forgeries.} 
MIMLv2 contains numerous modern images that were recently captured and forged, reflecting current photography and forgery technologies. In contrast, the CASIA dataset~\cite{casia} proposed over a decade ago, features blurred small images and outdated forgeries. Therefore, our MIMLv2 better suits current IML requirements.

\item \textbf{Strong Scalability.} 
With the rising popularity of online image manipulation competitions (e.g., PS-Battles~\cite{reddit}, which attracts over 20 million participants), new manually forged images are continuously generated.
Our dataset construction approach is ready to harness this growing cheap web data. Therefore, our dataset can be easily expanded in the future.
\end{itemize}

\begin{table}[t!]
\setlength{\tabcolsep}{1.2pt}
\caption{Comparison between different handcrafted image manipulation localization datasets. 'Num.' denotes number.}
\vspace{-0.1cm}
\begin{tabular}{ccccccccc}
\hline
Name        &  & Year &  & Real Num.             &  & Forged Num.            &  & (Height, Width) Range                         \\ \cline{1-1} \cline{3-3} \cline{5-5} \cline{7-7}  \cline{9-9} 
CASIAv1~\cite{casia}     &  & 2013 &  & 800             & & 921             &  & (246, 384)-(500, 334)          \\
CASIAv2~\cite{casia}     &  & 2013 &  & 7,461             & & 5,123            &  & (160, 240)-(901, 600)          \\
Coverage~\cite{coverage}    &  & 2016 &  & 100             & & 100             &  & (190, 334)-(472, 752)          \\
NIST16~\cite{nist16}     &  & 2016 &  & 224             & & 564             &  & (500, 500)-(3744, 5616)        \\
In Wild~\cite{inwild}        &  & 2018 &  & 0             &  & 201             &  & (650, 650)-(2736, 3648)        \\
MISD~\cite{misd}       &  & 2021 &  & 618             & & 300            &  & (160, 240)-(901, 600)        \\
IMD20~\cite{imd20}       &  & 2020 &  & 414             & & 2,010            &  & (193, 260)-(4437, 2958)        \\
CIMD~\cite{cimd}       &  & 2024 &  & 102             & & 102            &  & (2048, 1365)-(2048, 1365)        \\
MIMLv1~\cite{mimlv1} &  & 2024 &  & 0             & & 123,150 &  & \textbf{(45, 120)-(13846, 9200)} \\
\rowcolor{gray!15}MIMLv2 (Ours) &  & \textbf{2025} &  & \textbf{63,847}             & & \textbf{246,212} &  & \textbf{(45, 120)-(13846, 9200)} \\ \hline
\end{tabular}
\vspace{-0.1cm}
\label{tab: datas}
\end{table}

\section{Object Jitter}
\begin{figure*}[t]
 	\centering
 	\includegraphics[width=1.0\linewidth]{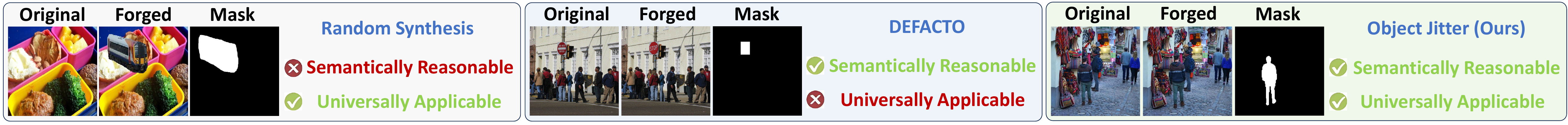}
 	\centering
 	\setlength{\abovecaptionskip}{-0.3cm}
 	\caption{Comparison between different data generation methods.
 	} \label{fig:Fig9_synthetic}
 \vspace{-0.4cm}
 \end{figure*}

Sophisticated copy-move forgeries, where an object is copied and pasted within the same image, are difficult to detect. They introduce only subtle edge artifacts while preserving source texture and noise patterns. Their scarcity in web-sourced data leads models to rely more on common noise and compression anomalies, causing them to fail on these hard cases. Existing synthesis methods, designed to generate training data for such forgeries, are flawed: random copy-paste techniques~\cite{liu2022pscc, zhou2018learning} often create semantically incoherent artifacts (e.g., a train on food, Fig.~\ref{fig:Fig9_synthetic}), while more constrained approaches like DEFACTO limit tampering to a few objects of consistent shape, lacking object diversity. Both issues can cause models to overfit to unrealistic cues.

To address this gap, we propose Object Jitter, a novel strategy that simulates these forgeries by subtly distorting existing objects. The core operation is Size Jitter: we slightly enlarge a randomly selected object, covering its original footprint. This mimics a realistic copy-move forgery by preserving semantics and texture while introducing only faint boundary artifacts. As illustrated in Fig. 9, the process involves (1) segmenting an object using ground truth or SAM~\cite{SAM}, (2) scaling the object and its mask by a small random factor, and (3) applying a smooth alpha-blended transition at the new boundary to minimize extraneous visual cues.

To further enhance data diversity, we introduce two additional minor perturbations: Exposure Jitter (slight overexposure) and Texture Jitter (subtle modifications via JPEG compression, reverse JPEG~\cite{fbcnn}, and blur operations). By preserving semantic plausibility and being applicable to any object, Object Jitter provides a source of challenging training examples. This compels the model to learn the subtle boundary artifacts characteristic of sophisticated forgeries, thereby improving its robustness and generalization.

\begin{figure}[t!]
 	\centering
 	\includegraphics[width=1.0\linewidth]{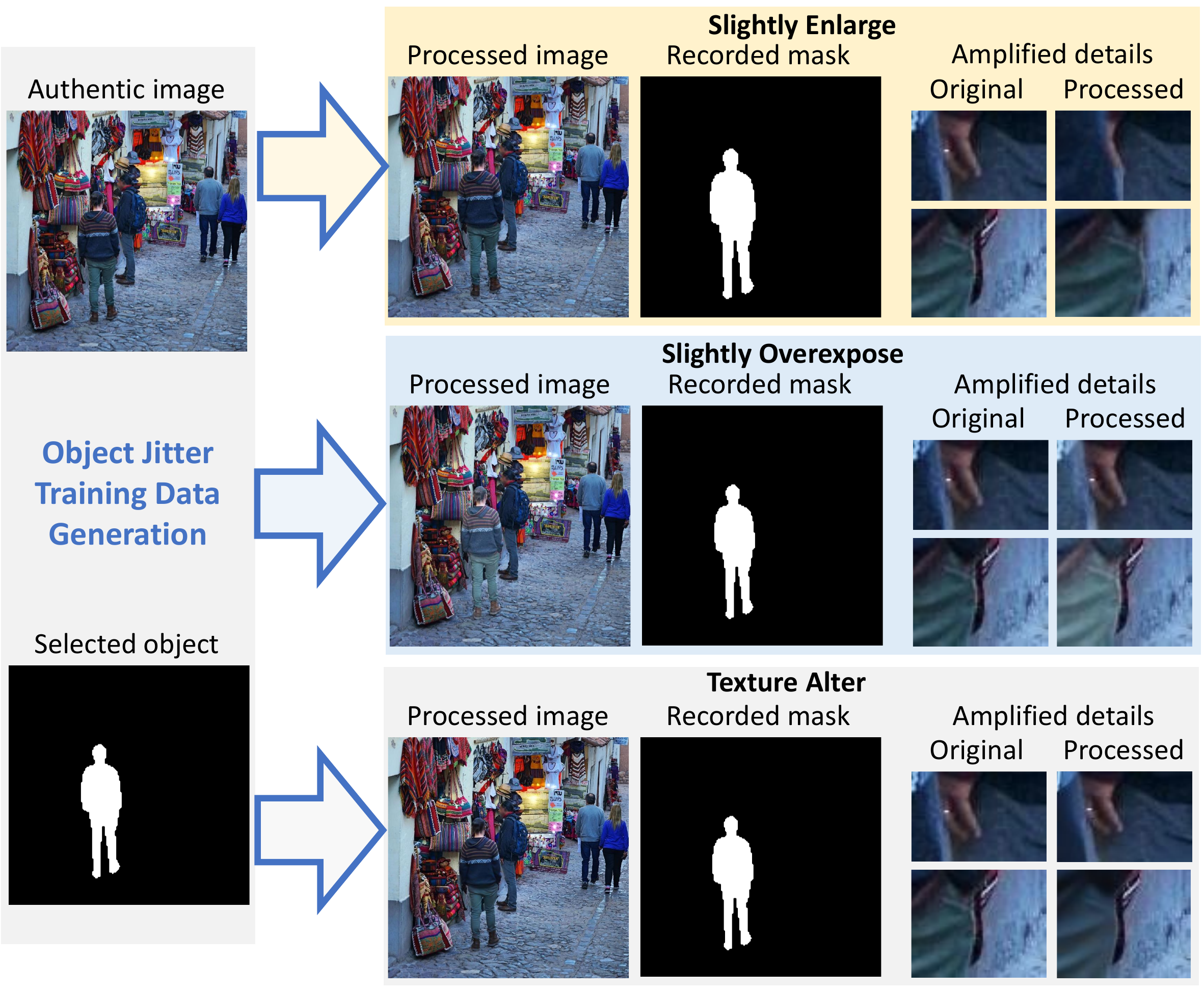}
 	\centering
 	\setlength{\abovecaptionskip}{-0.4cm}
 	\caption{The proposed Object Jitter method. Given a randomly selected object from an authentic image, we slightly enlarge or overexpose it, or alter its texture. The processed object is regarded as a tampered object for training. 
 	} \label{fig:Fig9_objectjt}
 \end{figure}

\section{Web-IML}
\begin{figure}[t!]
 	\centering
 	\includegraphics[width=1.0\linewidth]{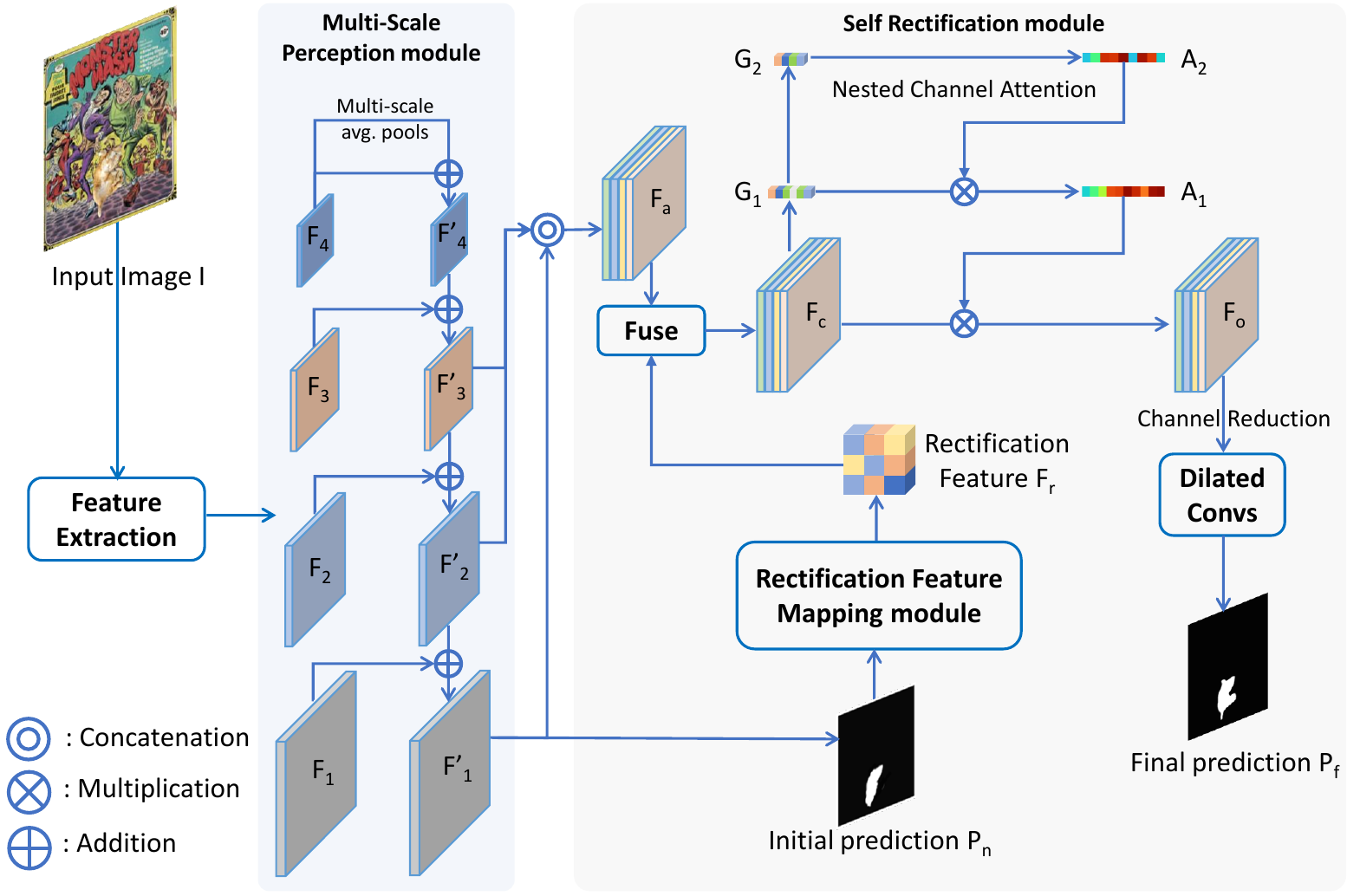}
 	\centering
 	\setlength{\abovecaptionskip}{-0.4cm}
 	\caption{The overall framework of the proposed Web-IML. 
 	} \label{fig:Fig10_WebNet}
 \end{figure}

To better leverage the web-scale supervision, we propose a new model called Web-IML. As illustrated in Fig.~\ref{fig:Fig10_WebNet}, Web-IML comprises a feature extractor, a Multi-Scale Perception module to integrate information from multiple perspectives, and a Self-Rectification module that enables the model to identify and correct its mistakes.

\subsection{Multi-Scale Perception Module}

\textbf{Motivation}. This module is designed to mimic the meticulous image forensic analysis, where humans often zoom in and out the image repeatedly, comparing diverse observations to assist their final prediction.

 \textbf{Pipeline}. Given the feature maps ${F_{1},F_{2},F_{3},F_{4}}$ extracted from the encoder, we first extract global features of different scales through multi-scale average pooling. The global features are then fused and dimension-reduced to get $F'4$. Next, the high-level features are fused with the low-level features recursively. This process fully integrates high-level semantics with low-level image details, resulting in the multi-scale features ${F'_{1},F'_{2},F'_{3},F'_{4}}$. More details are provided in the Appendix.

\subsection{Self-Rectification Module}

\textbf{Motivation}. This module is designed to make the model learn to check and correct its initial predictions, thereby minimizing the overlooking of key information and mistakes. The Self-Rectification (SR) module improves upon the Self-Calibration (SC) module in our conference paper from two aspects: First, in SC, we map the initial prediction mask to a convolution kernel, use it to convolve the mask and multiply the result by the multi-scale features. Despite its effectiveness, learning the map function to a convolution kernel is difficult, and directly multiplying the convolved mask often causes information loss in the black mask region. In our SR, we map the initial prediction mask to a feature map, and concatenate it with the multi-scale features, addressing the issues. Second, we introduce a Nested Channel Attention module that enables more meticulous analyses of the rectification feature.

\textbf{Pipeline}. As shown at the bottom of Fig.~\ref{fig:Fig10_WebNet}, we obtain the initial mask prediction $P_{n}$ from $F'_{1}$. We then adaptively obtain a rectification feature $F_{r}$ through a tiny Rectification Feature Mapping module ($RFM$), which consists of 5 cascaded conv-layers. The $F_{r}$ informs the model of suspected tampered regions and potentially incorrect prediction regions. $F_{r}$ is fused with $F_{a}$ to get $F_{c}$. Then, in the Nested Channel Attention module, we obtain the global vectors $G_{1}$ from $F_{c}$, and subsequently higher-level global vectors $G_{2}$ from $G_{1}$. Then $G_{1}$ is weighted by the attention map calculated by $G_{2}$ and the results are further used to calculate the channel attention weights for $F_{c}$.  The output $F_{o}$ is further processed by four dilated conv-layers, the output features are concatenated and fused to produce the mask prediction $P_{f}$. To get the refined final prediction $P_{f}$, we replace $P_{n}$ with $P_{f}$, replace $F_{a}$ with $F_{o}$, then repeat the self-rectification once more.

Our Web-IML is optimized with cross entropy loss in an end-to-end manner.

\begin{table*}[]
\setlength{\tabcolsep}{1pt}
\caption{Comparison on image manipulation localization. `WS` is our proposed web supervision, which trains models using data from the MIMLv2 and Object Jitter. `Avg*` is the average score excluding IMD20. `Param.` is the number of model parameters. CAT-Net, TruFor and SparseViT were trained with the IMD20 dataset and therefore not tested on it.}
\vspace{-0.1cm}
\begin{tabular}{ccccccccccccccccccccccccccccccccccccccc}
\hline
\multirow{2}{*}{Method} &  & \multicolumn{3}{c}{CASIAv1~\cite{casia}} &  & \multicolumn{3}{c}{Coverage~\cite{coverage}} &  & \multicolumn{3}{c}{NIST16~\cite{nist16}} &  & \multicolumn{3}{c}{IMD20~\cite{imd20}} &  & \multicolumn{3}{c}{CocoGlide~\cite{guillaro2023trufor}} &  & \multicolumn{3}{c}{CIMD~\cite{cimd}} &  & \multicolumn{3}{c}{MISD~\cite{misd}} &  & \multicolumn{3}{c}{Avg.*} &  & \multicolumn{3}{c}{Average} & & \multirow{2}{*}{Param.} \\ \cline{3-5} \cline{7-9} \cline{11-13} \cline{15-17} \cline{19-21} \cline{23-25} \cline{27-29} \cline{31-33} \cline{35-37} 
 & \multicolumn{1}{c}{} & \multicolumn{1}{c}{IoU} & \multicolumn{1}{c}{} & \multicolumn{1}{c}{F1} & \multicolumn{1}{c}{} & \multicolumn{1}{c}{IoU} & \multicolumn{1}{c}{} & \multicolumn{1}{c}{F1} & \multicolumn{1}{c}{} & \multicolumn{1}{c}{IoU} & \multicolumn{1}{c}{} & \multicolumn{1}{c}{F1} & \multicolumn{1}{c}{} & \multicolumn{1}{c}{IoU} & \multicolumn{1}{c}{} & \multicolumn{1}{c}{F1} & \multicolumn{1}{c}{} & \multicolumn{1}{c}{IoU} & \multicolumn{1}{c}{} & \multicolumn{1}{c}{F1} & \multicolumn{1}{c}{} & \multicolumn{1}{c}{IoU} & \multicolumn{1}{c}{} & \multicolumn{1}{c}{F1} & \multicolumn{1}{c}{} & \multicolumn{1}{c}{IoU} & \multicolumn{1}{c}{} & \multicolumn{1}{c}{F1} & \multicolumn{1}{c}{} & \multicolumn{1}{c}{IoU} & \multicolumn{1}{c}{} & \multicolumn{1}{c}{F1} & \multicolumn{1}{c}{} & \multicolumn{1}{c}{IoU} & \multicolumn{1}{c}{} & \multicolumn{1}{c}{F1} & \\ \cline{1-1} \cline{3-3} \cline{5-5} \cline{7-7} \cline{9-9} \cline{11-11} \cline{13-13} \cline{15-15} \cline{17-17} \cline{19-19} \cline{21-21} \cline{23-23} \cline{25-25} \cline{27-27} \cline{29-29} \cline{31-31} \cline{33-33} \cline{35-35} \cline{37-37}  \cline{39-39} 
ManTra-Net~\cite{wu2019mantra} &  & .086 &  & .130 &  & .181 &  & .271 &  & .040 &  & .062 &  & .098 &  & .146 &  & .155 &  & .203 &  & .058 &  & .096 &  & .105 &  & .171 &  & .104 &  & .156 &  & .103 &  & .154  &  & 8M \\
RRU-Net~\cite{rru} &  & .330 &  & .380 &  & .165 &  & .260 &  & .080 &  & .129 &  & .169 &  & .256 &  & .223 &  & .304 &  & .036 &  & .068 &  & .440 &  & .570 &  & .212 &  & .285 &  & .206 &  & .281  &  & 5M \\
MVSS-Net~\cite{dong2022mvss} &  & .403 &  & .435 &  & .389 &  & .454 &  & .243 &  & .294 &  & .200 &  & .260 &  & .276 &  & .357 &  & .016 &  & .021 &  & .120 &  & .175 &  & .235 &  & .284 &  & .235 &  & .285  &  & 146M \\
PSCC-Net~\cite{liu2022pscc} &  & .410 &  & .463 &  & .340 &  & .446 &  & .067 &  & .110 &  & .115 &  & .192 &  & .333 &  & .422 &  & .212 &  & .290 &  & .400 &  & .491 &  & .294 &  & .370 &  & .268 &  & .345  &  & \textbf{4M} \\
CAT-Net~\cite{catnet} &  & .684 &  & .738 &  & .238 &  & .292 &  & .238 &  & .302 &  & - &  & - &  & .290 &  & .366 &  & .282 &  & .344 &  & .314 &  & .394 &  & .341 &  & .406 &  & - &  & -  &  & 114M \\
IF-OSN~\cite{osn} &  & .465 &  & .509 &  & .181 &  & .268 &  & .247 &  & .326 &  & .259 &  & .364 &  & .207 &  & .264 &  & .195 &  & .295 &  & .375 &  & .464 &  & .278 &  & .354 &  & .341 &  & .406  &  & 124M \\
EVP~\cite{evp} &  & .438 &  & .502 &  & .078 &  & .114 &  & .188 &  & .239 &  & .177 &  & .268 &  & .084 &  & .118 &  & .224 &  & .283 &  & .392 &  & .531 &  & .234 &  & .298 &  & .276 &  & .356  &  & 63M \\
TruFor~\cite{guillaro2023trufor} &  & .630 &  & .692 &  & .446 &  & .522 &  & .279 &  & .348 &  & - &  & - &  & .294 &  & .362 &  & .322 &  & .418 &  & .426 &  & .549 &  & .400 &  & .482 &  & - &  & -  &  & 67M \\
SparseViT~\cite{sparsevit} &  & .768 &  & .819 &  & .456 &  & .504 &  & .322 &  & .381 &  & - &  & - &  & .327 &  & .389 &  & .257 &  & .322 &  & .421 &  & .538 &  & .425 &  & .492 &  & - &  & -  &  & 50M \\
PIM~\cite{pami} &  & .512 &  & .566 &  & .188 &  & .251 &  & .225 &  & .280 &  & .340 &  & .419 &  & .327 &  & .404 &  & .177 &  & .229 &  & .389 &  & .519 &  & .303 &  & .375 &  & .308 &  & .381 &  & 168M \\
\hline
APSC-Net~\cite{mimlv1} (\textit{Conf.}) &  & .799 &  & .837 &  & .490 &  & .523 &  & .398 &  & .436 &  & .339 &  & .391 &  & .243 &  & .283 &  & .517 &  & .565 &  & .420 &  & .524 &  & .477 &  & .528 &  & .458 &  & .508  &  & 143M \\
\rowcolor{gray!15} Web-IML (\textit{Ours}) &  & .823 &  & .857 &  & .508 &  & .551 &  & .412 &  & .465 &  & .370 &  & .422 &  & .285 &  & .339 &  & .629 &  & .673 &  & .430 &  & .545 &  & .515 &  & .572 &  & .494 &  & .550  &  & 140M \\
APSC-Net (w/ MIMLv1) &  & .810 &  & .848 &  & .498 &  & .568 &  & .525 &  & .590 &  & .679 &  & .760 &  & .292 &  & .355 &  & .652 &  & .720 &  & .484 &  & .587 &  & .543 &  & .611 &  & .562 &  & .632  &  & 143M \\
\rowcolor{gray!15} Web-IML (w/ MIMLv1) &  & .821 &  & .854 &  & .561 &  & .616 &  & .520 &  & .587 &  & .692 &  & .768 &  & .275 &  & .325 &  & \textbf{.789} &  & \textbf{.845} &  & .496 &  & .610 &  & .578 &  & .640 &  & .594 &  & .658  &  & 140M \\
APSC-Net (w/ WS) &  & .832 &  & .865 &  & .594 &  & .642 &  & .578 &  & .661 &  & .690 &  & .766 &  & .389 &  & .451 &  & .660 &  & .725 &  & .485 &  & .588 &  & .598 &  & .655 &  & .604 &  & .671  &  & 143M \\
\rowcolor{gray!15} Web-IML (w/ WS) &  & \textbf{.846} & \textbf{} & \textbf{.879} & \textbf{} & \textbf{.715} & \textbf{} & \textbf{.768} & \textbf{} & \textbf{.591} & \textbf{} & \textbf{.670} & \textbf{} & \textbf{.698} & \textbf{} & \textbf{.776} & \textbf{} & \textbf{.439} & \textbf{} & \textbf{.507} & \textbf{} & .759 & \textbf{} & .821 & \textbf{} & \textbf{.495} & \textbf{} & \textbf{.612} & \textbf{} & \textbf{.641} & \textbf{} & \textbf{.709} & \textbf{} & \textbf{.649} & \textbf{} & \textbf{.719}  &  & 140M \\ \hline
\end{tabular}
\label{tab: maincomp}
\end{table*}

\begin{table*}[]
\setlength{\tabcolsep}{2pt}
\caption{Ablation study on our Web-IML model. `MP` is the Multi-Scale Perception module. `SR` is the Self-Rectification module. `NA` is our Nested Channel Attention. `RF` denotes the second round self-rectification. `WS` denotes the proposed Web Supervision.}
\vspace{-0.1cm}
\begin{tabular}{ccccccccccccccccccccccccccccccccccccccccccc}
\hline
\multirow{2}{*}{Num.} &  & \multicolumn{9}{c}{Ablation Settings} &  & \multicolumn{3}{c}{CASIAv1~\cite{casia}} &  & \multicolumn{3}{c}{Coverage~\cite{coverage}} &  & \multicolumn{3}{c}{NIST16~\cite{nist16}} &  & \multicolumn{3}{c}{IMD20~\cite{imd20}} &  & \multicolumn{3}{c}{CocoGlide~\cite{guillaro2023trufor}} &  & \multicolumn{3}{c}{CIMD~\cite{cimd}} &  & \multicolumn{3}{c}{MISD~\cite{misd}} &  & \multicolumn{3}{c}{Average} \\ \cline{3-11} \cline{13-15} \cline{17-19} \cline{21-23} \cline{25-27} \cline{29-31} \cline{33-35} \cline{37-39} \cline{41-43} 
 &  & MP &  & SR & & NA & & RF &  & WS &  & IoU &  & F1 &  & IoU &  & F1 &  & IoU &  & F1 &  & IoU &  & F1 &  & IoU &  & F1 &  & IoU &  & F1 &  & IoU &  & F1 &  & IoU &  & F1 \\ \cline{1-1} \cline{3-3} \cline{5-5} \cline{7-7} \cline{9-9} \cline{11-11} \cline{13-13} \cline{15-15} \cline{17-17} \cline{19-19} \cline{21-21} \cline{23-23} \cline{25-25} \cline{27-27} \cline{29-29} \cline{31-31} \cline{33-33} \cline{35-35} \cline{37-37} \cline{39-39} \cline{41-41} \cline{43-43}
(1) &  & $\times$ &  & $\times$ &  & $\times$ &  & $\times$ &  & $\times$ &  & .711 &  & .779 &  & .361 &  & .430 &  & .346 &  & .410 &  & .273 &  & .342 &  & .345 &  & .417 &  & .262 &  & .338 &  & .362 &  & .471 &  & .380 &  & .455 \\
(2) &  & \checkmark &  & $\times$ &  & $\times$ &  & $\times$ &  & $\times$ &  & .775 &  & .814 &  & .440 &  & .487 &  & .359 &  & .415 &  & .351 &  & .406 &  & .275 &  & .324 &  & .450 &  & .502 &  & .375 &  & .482 &  & .432 &  & .490 \\
(3) &  & \checkmark &  & \checkmark &  & $\times$ &  & $\times$ &  & $\times$ &  & \multicolumn{1}{c}{.802} &  & \multicolumn{1}{c}{.836} &  & \multicolumn{1}{c}{.523} & \textbf{} & \multicolumn{1}{c}{.576} &  & \multicolumn{1}{c}{.385} &  & \multicolumn{1}{c}{.441} &  & \multicolumn{1}{c}{.364} &  & \multicolumn{1}{c}{.415} &  & \multicolumn{1}{c}{.287} & \textbf{} & \multicolumn{1}{c}{.343} &  & \multicolumn{1}{c}{.557} &  & \multicolumn{1}{c}{.610} &  & \multicolumn{1}{c}{.382} &  & \multicolumn{1}{c}{.493} &  & \multicolumn{1}{c}{.471} &  & \multicolumn{1}{c}{.530} \\
(4) &  & \checkmark &  & \checkmark &  & \checkmark &  & $\times$ &  & $\times$ &  & \multicolumn{1}{c}{.816} &  & \multicolumn{1}{c}{.848} &  & \multicolumn{1}{c}{\textbf{.549}} & \textbf{} & \multicolumn{1}{c}{\textbf{.598}} &  & \multicolumn{1}{c}{.394} &  & \multicolumn{1}{c}{.449} &  & \multicolumn{1}{c}{.369} &  & \multicolumn{1}{c}{.419} &  & \multicolumn{1}{c}{\textbf{.291}} & \textbf{} & \multicolumn{1}{c}{\textbf{.345}} &  & \multicolumn{1}{c}{.593} &  & \multicolumn{1}{c}{.649} &  & \multicolumn{1}{c}{.386} &  & \multicolumn{1}{c}{.496} &  & \multicolumn{1}{c}{.486} &  & \multicolumn{1}{c}{.543} \\
(5) &  & \checkmark &  & \checkmark &  & \checkmark &  & \checkmark &  & $\times$ &  & \textbf{.823} & \textbf{} & \textbf{.857} & \textbf{} & .508 &  & .551 & \textbf{} & \textbf{.412} & \textbf{} & \textbf{.465} &  & \textbf{.370} & \textbf{} & \textbf{.422} &  & .285 &  & .339 & \textbf{} & \textbf{.629} & \textbf{} & \textbf{.673} & \textbf{} & \textbf{.430} & \textbf{} & \textbf{.545} & \textbf{} & \textbf{.494} & \textbf{} & \textbf{.550} \\ \hline
(6) &  & \checkmark &  & \checkmark &  & \checkmark &  & \checkmark &  & \checkmark &  & \textbf{.846} & \textbf{} & \textbf{.879} &  & \textbf{.715} &  & \textbf{.768} &  & \textbf{.591} &  & \textbf{.670} &  & \textbf{.698} &  & \textbf{.776} &  & \textbf{.439} &  & \textbf{.507} & \textbf{} & \textbf{.759} & \textbf{} & \textbf{.821} &  & \textbf{.496} &  & \textbf{.612} & \textbf{} & \textbf{.649} & \textbf{} & \textbf{.719} \\ \hline
\end{tabular}
\label{tab: modelabl}
\end{table*}

\begin{table}[t!]
\setlength{\tabcolsep}{3pt}
\caption{Robustness evaluation with AUC metric on NIST16 dataset.}
\vspace{-0.1cm}
\begin{tabular}{cccccccccccc}
\hline
                         &  &                            &  & \multicolumn{2}{c}{Resize}    &  & \multicolumn{2}{c}{{Blur}} &           & \multicolumn{2}{c}{{JPEG}} \\ \cline{5-6} \cline{8-9} \cline{11-12} 
\multirow{-2}{*}{Method} &  & \multirow{-2}{*}{Ori} &  & .78x         & .25x         &  & k=3                                    & k=15            &           & q=100                                    & q=50             \\ \cline{1-1} \cline{3-3} \cline{5-6} \cline{8-9} \cline{11-12} 
ManTra-Net~\cite{wu2019mantra}               &  & .795                       &  & .774          & .755          &  & {.774}            & .746            &           & {.779}              & .744             \\
SPAN~\cite{hu2020span}                     &  & .840                       &  & .832          & .802          &  & {.831}            & .792            &           & {.836}              & .807             \\
PSCC-Net~\cite{liu2022pscc}                &  & .855                       &  & .853          & .850          &  & {.854}            & .800            &           & {.854}              & .854             \\
ObjectFormer~\cite{wang2022objectformer}             &  & .872                       &  & .872          & .863          &  & .860                                   & .803            &           & .864                                     & .862             \\
SparseViT~\cite{sparsevit}                &  & .888                       &  & .884          & .869          &  & .881                                   & .877            &           & .886                                     & .881             \\
NCL~\cite{ncliml}                      &  & .912                       &  & .856          & .831          &  & .840                                   & .806            &           & .843                                     & .819             \\
ERMPC~\cite{ermpc}                       &  & .895                       &  & .893         & .877          &  & .892                                   & .871           &           & .894                                     & .888             \\
UnionFormer~\cite{li2024unionformer}                       &  & .881                      &  & .873         & .872          &  & .865                                   & .843           &           & .880                                     & .880             \\
\hline
APSC-Net+MIMLv1                 &  & .928                       &  & .917          & .888          &  & {.907}            & .900            &           & {.922}              & .907             \\
\rowcolor{gray!15}Web-IML (Ours)          &  & \textbf{.942}              &  & \textbf{.935} & \textbf{.904} &  & {\textbf{.914}}   & \textbf{.900}   & \textbf{} & {\textbf{.940}}     & \textbf{.909}    \\ \hline
\end{tabular}
\vspace{-0.2cm}
\label{tab: robust}
\end{table}

\begin{table*}[]
\setlength{\tabcolsep}{1.2pt}
\caption{Ablation study on our Web Supervision method. `PSCC-Synthetic` denotes the synthetic data from the PSCC-Net~\cite{liu2022pscc}. `O.J.` denotes the proposed Object Jitter method. `C.G.` denotes the CocoGlide~\cite{guillaro2023trufor}. `$\Delta$` denotes improvement.}
\begin{tabular}{ccccccccccccccccccccccccccccccccccc}
\hline
& & &  & \multicolumn{3}{c}{CASIAv1~\cite{casia}} &  & \multicolumn{3}{c}{Coverage~\cite{coverage}} &  & \multicolumn{3}{c}{NIST16~\cite{nist16}} &  & \multicolumn{3}{c}{IMD20~\cite{imd20}} &  & \multicolumn{3}{c}{C.G.~\cite{guillaro2023trufor}} &  & \multicolumn{3}{c}{CIMD~\cite{cimd}} &  & \multicolumn{3}{c}{MISD~\cite{misd}} &  & \multicolumn{3}{c}{Average (\#$\Delta$)} \\ \cline{5-7} \cline{9-11} \cline{13-15} \cline{17-19} \cline{21-23} \cline{25-27} \cline{29-31} \cline{33-35} 
\multirow{-2}{*}{Num.}& & \multirow{-2}{*}{Training Data} &  & IoU &  & F1 &  & IoU &  & F1 &  & IoU &  & F1 &  & IoU &  & F1 &  & IoU &  & F1 &  & IoU &  & F1 &  & IoU &  & F1 &  & IoU &  & F1 \\ \cline{1-1} \cline{3-3} \cline{5-5} \cline{7-7} \cline{9-9} \cline{11-11} \cline{13-13} \cline{15-15} \cline{17-17} \cline{19-19} \cline{21- 21} \cline{23-23} \cline{25-25} \cline{27-27} \cline{29-29} \cline{31-31} \cline{33-33}  \cline{35-35} 
(1) & & Web-IML Baseline &  & .823 &  & .857 &  & .508 &  & .551 &  & .412 &  & .465 &  & .370 &  & .422 &  & .285 &  & .339 &  & .629 &  & .673 &  & .430 &  & .545 &  & .494 &  & .550 \\
(2) & & +PSCC-Synthetic~\cite{liu2022pscc} &  & .813 &  & .845 &  & .503 &  & .546 &  & .348 &  & .392 &  & .361 &  & .410 &  & .177 &  & .211 &  & .598 &  & .654 &  & .330 &  & .441 &  & .447 (-9.5\%) &  & .500 (-9.1\%) \\
(3) & & +DEFACTO~\cite{defacto} &  & .809 &  & .843 &  & .354 &  & .380 &  & .412 &  & .465 &  & .365 &  & .424 &  & .359 &  & .418 &  & .394 &  & .438 &  & .461 &  & .572 &  & .451 (-8.7\%) &  & .506 (-8.0\%) \\
(4) & & +AIGC-Synthetic~\cite{wang2025opensdi} &  & .817 &  & .849 &  & .273 &  & .296 &  & .220 &  & .276 &  & .284 &  & .337 &  & .394 &  & .462 &  & .287 &  & .340 &  & .373 &  & .429 &  & .376 (-24.\%) &  & .420 (-23.\%) \\
(5) & & +WSCL~\cite{wscl} &  & .815 &  & .849 &  & .467 &  & .510 &  & .415 &  & .468 &  & .411 &  & .464 &  & .293 &  & .347 &  & .630 &  & .672 &  & .431 & & .544 &  & .495 (+0.2\%) & & .551 (+0.2\%) \\ \hline
(6) & & +MIMLv1~\cite{mimlv1} (\textit{Conf.}) &  & .821 &  & .854 &  & .561 &  & .616 &  & .520 &  & .587 &  & .692 &  & .768 &  & .275 &  & .325 &  & .789 &  & .845 &  & .496 &  & .610 &  & .594 (+20.\%) &  & .658 (+20.\%) \\
\rowcolor{gray!15}(7) & &  +MIMLv2 &  & \textbf{.849} &  & \textbf{.883} &  & .694 &  & .755 &  & .518 &  & .584 &  & .697 &  & .772 &  & .376 &  & .435 &  & .716 &  & .776 &  & .448 &  & .564 &  & .614 (+24.\%) &  & .681 (+24.\%) \\
\rowcolor{gray!15}(8) & &  +MIMLv1~\cite{mimlv1}+O.J. &  & .833 &  & .867 &  & .587 &  & .643 &  & .549 &  & .620 &  & .693 &  & .768 &  & .344 &  & .407 &  & .793 &  & .849 &  & .487 &  & .602 &  & .612 (+24.\%) &  & .679 (+23.\%) \\
\rowcolor{gray!15}(9) & &  +O.J. &  & .822 &  & .853 &  & .575 &  & .616 &  & .270 &  & .310 &  & .422 &  & .482 &  & .308 &  & .362 &  & .698 &  & .760 &  & .413 &  & .521 &  & .501 (+1.4\%) &  & .558 (+1.4\%) \\
\rowcolor{gray!15}(10) & &  +MIMLv2+O.J. (Ours) &  & .846 &  & .879 &  & \textbf{.715} &  & \textbf{.768} &  & \textbf{.591} &  & \textbf{.670} &  & .698 &  & \textbf{.776} &  & .439 &  & .507 &  & .759 &  & .821 &  & \textbf{.496} &  & .612 &  & \textbf{.649 (+31.\%)}  &  & \textbf{.719 (+31.\%)} \\ \hline
\end{tabular}
\label{tab: dataabl}
\end{table*}

\begin{table*}[]
\setlength{\tabcolsep}{1.8pt}
\caption{Ablation study on our Object Jitter method. `COCO` denotes using the COCO~\cite{lin2014microsoft} images for Object Jitter. `SA` denotes using 400k random images from the SA-1B~\cite{SAM} dataset for Object Jitter. `C.G.` denotes the CocoGlide dataset~\cite{guillaro2023trufor}.}
\begin{tabular}{ccccccccccccccccccccccccccccccccccccccccc}
\hline
\multirow{2}{*}{Num} &  & \multicolumn{5}{c}{Object Jitter Operation} &  & \multirow{2}{*}{\begin{tabular}[c]{@{}c@{}}Image\\ Source\end{tabular}} &  & \multicolumn{3}{c}{CASIAv1~\cite{casia}} &  & \multicolumn{3}{c}{Coverage~\cite{coverage}} &  & \multicolumn{3}{c}{NIST16~\cite{nist16}} &  & \multicolumn{3}{c}{IMD20~\cite{imd20}} &  & \multicolumn{3}{c}{C.G.~\cite{guillaro2023trufor}} &  & \multicolumn{3}{c}{CIMD~\cite{cimd}} &  & \multicolumn{3}{c}{MISD~\cite{misd}} &  & \multicolumn{3}{c}{Average} \\ \cline{3-7} \cline{11-13} \cline{15-17} \cline{19-21} \cline{23-25} \cline{27-29} \cline{31-33} \cline{35-37} \cline{39-41} 
 &  & Size  &  & Texture &  & Exposure &  &  &  & IoU &  & F1 &  & IoU &  & F1 &  & IoU &  & F1 &  & IoU &  & F1 &  & IoU &  & F1 &  & IoU &  & F1 &  & IoU &  & F1 &  & IoU &  & F1 \\ \cline{1-1} \cline{3-3} \cline{5-5} \cline{7-7} \cline{9-9} \cline{11-11} \cline{13-13} \cline{15-15} \cline{17-17} \cline{19-19} \cline{21-21} \cline{23-23} \cline{25-25} \cline{27-27} \cline{29-29} \cline{31-31} \cline{33-33} \cline{35-35} \cline{37-37} \cline{39-39} \cline{41-41} 
(1) &  & $\times$ &  & $\times$ &  & $\times$ &  & SA &  & \textbf{.849} & \textbf{} & \textbf{.883} &  & .694 &  & .755 &  & .518 &  & .584 &  & .697 &  & .772 &  & .376 &  & .435 &  & .716 &  & .776 &  & .448 &  & .564 &  & .614 &  & .681 \\
(2) &  & $\times$ &  & \checkmark &  & \checkmark &  & SA &  & .839 &  & .872 &  & .634 &  & .687 &  & .528 &  & .592 &  & \textbf{.703} & \textbf{} & \textbf{.778} &  & .396 &  & .456 &  & .748 &  & .798 &  & .487 &  & .605 &  & .619 &  & .684 \\
(3) &  & \checkmark &  & $\times$ &  & \checkmark &  & SA &  & .844 &  & .878 &  & .669 &  & .729 &  & .554 &  & .618 &  & .701 &  & .778 &  & .427 &  & .491 &  & .780 &  & .834 &  & .501 &  & .616 &  & .640 &  & .706 \\
(4) &  & \checkmark &  & \checkmark &  & $\times$ &  & SA &  & .839 &  & .872 &  & .699 &  & .754 &  & .560 &  & .627 &  & .699 &  & .774 &  & .413 &  & .479 &  & .802 &  & .861 &  & .512 &  & .628 &  & .646 &  & .714 \\
(5) &  & \checkmark &  & $\times$ &  & $\times$ &  & SA &  & .841 &  & .875 &  & .673 &  & .732 &  & .540 &  & .602 &  & .701 &  & .777 &  & .432 &  & .496 &  & .758 &  & .816 &  & .516 &  & .631 &  & .637 &  & .704 \\
(6) &  & \checkmark &  & \checkmark &  & \checkmark &  & COCO &  & .848 &  & .881 &  & .645 &  & .711 &  & .584 &  & .657 &  & .699 &  & .776 &  & \textbf{.489} & \textbf{} & \textbf{.549} &  & \textbf{.803} & \textbf{} & \textbf{.864} &  & .496 &  & .612 &  & \textbf{.652} & \textbf{} & \textbf{.721} \\
(7) &  & \checkmark &  & \checkmark &  & \checkmark &  & SA &  & .846 &  & .879 &  & \textbf{.715} & \textbf{} & \textbf{.768} &  & \textbf{.591} & \textbf{} & \textbf{.670} &  & .698 &  & .776 &  & .439 &  & .507 &  & .759 &  & .821 &  & \textbf{.496} & \textbf{} & \textbf{.613} &  & .649 &  & .719 \\ \hline
\end{tabular}
\label{tab: ojabl}
\end{table*}

\begin{table*}[]
\setlength{\tabcolsep}{0.55pt}
\caption{Comparison study of our web-supervision on IML task. \textcolor{blue}{Blue number for improvement} and \textcolor{red}{Red number for decrease}. `B.L.` denotes baseline setting. `PSCC-Synth.` denotes the synthetic dataset of PSCC-Net~\cite{liu2022pscc}, `AIGC-Synth.` denotes the AIGC forgery dataset OpenSDID~\cite{wang2025opensdi}. `O.J.` denotes our Object Jitter method.}
\begin{tabular}{ccccccccccccccccccccccccccccccccccccccccccccccccc}
\hline
\multirow{2}{*}{Training Data} &  & \multicolumn{7}{c}{CASIAv1~\cite{casia}} &  & \multicolumn{7}{c}{Coverage~\cite{coverage}} &  & \multicolumn{7}{c}{NIST16~\cite{nist16}} &  & \multicolumn{7}{c}{IMD20~\cite{imd20}} &  & \multicolumn{7}{c}{CIMD~\cite{cimd}} &  & \multicolumn{7}{c}{Average} \\ \cline{3-9} \cline{11-17} \cline{19-25} \cline{27-33} \cline{35-41} \cline{43-49} 
 &  & IoU &  & $\Delta$ &  & F1 &  & $\Delta$ &  & IoU &  & $\Delta$ &  & F1 &  & $\Delta$ &  & IoU &  & $\Delta$ &  & F1 &  & $\Delta$ &  & IoU &  & $\Delta$ &  & F1 &  & $\Delta$ &  & IoU &  & $\Delta$ &  & F1 &  & $\Delta$ &  & IoU &  & $\Delta$ &  & F1 &  & $\Delta$ \\ \cline{1-1} \cline{3-3} \cline{5-5} \cline{7-7} \cline{9-9} \cline{11-11} \cline{13-13} \cline{15-15} \cline{17-17} \cline{19-19} \cline{21-21} \cline{23-23} \cline{25-25} \cline{27-27} \cline{29-29} \cline{31-31} \cline{33-33} \cline{35-35} \cline{37-37} \cline{39-39} \cline{41-41} \cline{43-43} \cline{45-45} \cline{47-47} \cline{49-49} 
PSCC-Net B.L. &  & .401 &  & 0 &  & .430 &  & 0 &  & .197 &  & 0 &  & .218 &  & 0 &  & .247 &  & 0 &  & .295 &  & 0 &  & .125 &  & 0 &  & .156 &  & 0 &  & .161 &  & 0 &  & .221 &  & 0 &  & .226 &  & 0 &  & .264 &  & 0 \\
+PSCC-Synth. &  & .352 &  & \textcolor{red}{12\%} &  & .379 &  & \textcolor{red}{12\%} &  & .147 &  & \textcolor{red}{25\%} &  & .159 &  & \textcolor{red}{27\%} &  & .230 &  & \textcolor{red}{7\%} &  & .274 &  & \textcolor{red}{7\%} &  & .101 &  & \textcolor{red}{19\%} &  & .125 &  & \textcolor{red}{20\%} &  & .101 &  & \textcolor{red}{37\%} &  & .124 &  & \textcolor{red}{44\%} &  & .186 &  & \textcolor{red}{18\%} &  & .212 &  & \textcolor{red}{20\%} \\
+DEFACTO\cite{defacto} &  & .394 &  & \textcolor{red}{2\%} &  & .429 &  & \textcolor{red}{1\%} &  & .231 &  & \textcolor{blue}{17\%} &  & .256 &  & \textcolor{blue}{17\%} &  & .223 &  & \textcolor{red}{10\%} &  & .270 &  & \textcolor{red}{8\%} &  & .157 &  & \textcolor{blue}{26\%} &  & .171 &  & \textcolor{blue}{10\%} &  & \textcolor{red}{.120} &  & \textcolor{red}{25\%} &  & .159 &  & \textcolor{red}{28\%} &  & .225 &  & \textcolor{red}{0\%} &  & .257 &  & \textcolor{red}{3\%} \\
+AIGC-Synth. &  & .346 &  & \textcolor{red}{16\%} &  & .377 &  & \textcolor{red}{14\%} &  & .109 &  & \textcolor{red}{81\%} &  & .123 &  & \textcolor{red}{77\%} &  & .091 &  & \textcolor{red}{171\%} &  & .142 &  & \textcolor{red}{108\%} &  & .110 &  & \textcolor{red}{14\%} &  & .144 &  & \textcolor{red}{8\%} &  & .124 &  & \textcolor{red}{30\%} &  & .170 &  & \textcolor{red}{30\%} &  & .156 &  & \textcolor{red}{45\%} &  & .191 &  & \textcolor{red}{38\%} \\
+WSCL~\cite{wscl} &  & .439 &  & \textcolor{blue}{9\%} &  & .464 &  & \textcolor{blue}{8\%} &  & .201 &  & \textcolor{blue}{2\%} &  & .220 &  & \textcolor{blue}{1\%} &  & .253 &  & \textcolor{blue}{2\%} &  & .300 &  & \textcolor{blue}{2\%} &  & .192 &  & \textcolor{blue}{54\%} &  & .235 &  & \textcolor{blue}{51\%} &  & .177 &  & \textcolor{blue}{10\%} &  & .239 &  & \textcolor{blue}{8\%} &  & .252 &  & \textcolor{blue}{12\%} &  & .293 &  & \textcolor{blue}{11\%} \\
+MIMLv1 (\textit{Conf.}) &  & .609 &  & \textcolor{blue}{52\%} &  & .649 &  & \textcolor{blue}{51\%} &  & .395 &  & \textcolor{blue}{101\%} &  & .477 &  & \textcolor{blue}{119\%} &  & .402 &  & \textcolor{blue}{63\%} &  & .476 &  & \textcolor{blue}{61\%} &  & .470 &  & \textcolor{blue}{276\%} &  & .541 &  & \textcolor{blue}{247\%} &  & .206 &  & \textcolor{blue}{28\%} &  & .251 &  & \textcolor{blue}{14\%} &  & .416 &  & \textcolor{blue}{84\%} &  & .479 &  & \textcolor{blue}{81\%} \\
\rowcolor{gray!15} +MIMLv2 &  & .602 &  & \textcolor{blue}{50\%} &  & .642 &  & \textcolor{blue}{49\%} &  & .402 &  & \textcolor{blue}{104\%} &  & .498 &  & \textcolor{blue}{128\%} &  & .405 &  & \textcolor{blue}{64\%} &  & .485 &  & \textcolor{blue}{64\%} &  & \textbf{.575} &  & \textbf{\textcolor{blue}{360\%}} &  & \textbf{.662} &  & \textbf{\textcolor{blue}{324\%}} &  & \textbf{.354} &  & \textbf{\textcolor{blue}{120\%}} &  & \textbf{.483} &  & \textbf{\textcolor{blue}{119\%}} &  & .468 &  & \textbf{\textcolor{blue}{107\%}} &  & .554 &  & \textcolor{blue}{110\%} \\
\rowcolor{gray!15} +MIMLv2+O.J &  & \textbf{.609} &  & \textbf{\textcolor{blue}{52\%}} &  & \textbf{.651} &  & \textbf{\textcolor{blue}{51\%}} &  & \textbf{.451} &  & \textbf{\textcolor{blue}{129\%}} &  & \textbf{.550} &  & \textbf{\textcolor{blue}{152\%}} &  & \textbf{.406} &  & \textbf{\textcolor{blue}{64\%}} &  & \textbf{.486} &  & \textbf{\textcolor{blue}{65\%}} &  & .570 &  & \textcolor{blue}{356\%} &  & .658 &  & \textcolor{blue}{322\%} &  & .335 &  & \textcolor{blue}{108\%} &  & .464 &  & \textcolor{blue}{110\%} &  & \textbf{.474} &  & \textbf{\textcolor{blue}{110\%}} &  & \textbf{.562} &  & \textbf{\textcolor{blue}{113\%}} \\ \hline
CAT-Net B.L. &  & .660 &  & 0 &  & .703 &  & 0 &  & .245 &  & 0 &  & .286 &  & 0 &  & .239 &  & 0 &  & .287 &  & 0 &  & .157 &  & 0 &  & .192 &  & 0 &  & .229 &  & 0 &  & .263 &  & 0 &  & .306 &  & 0 &  & .346 &  & 0 \\
+PSCC-Synth. &  & .410 &  & \textcolor{red}{38\%} &  & .437 &  & \textcolor{red}{38\%} &  & .215 &  & \textcolor{red}{12\%} &  & .232 &  & \textcolor{red}{19\%} &  & .288 &  & \textcolor{blue}{21\%} &  & .331 &  & \textcolor{blue}{15\%} &  & .121 &  & \textcolor{red}{23\%} &  & .148 &  & \textcolor{red}{23\%} &  & .160 &  & \textcolor{red}{30\%} &  & .189 &  & \textcolor{red}{28\%} &  & .239 &  & \textcolor{red}{22\%} &  & .268 &  & \textcolor{red}{23\%} \\
+DEFACTO\cite{defacto} &  & .673 &  & \textcolor{blue}{2\%} &  & .715 &  & \textcolor{blue}{2\%} &  & .200 &  & \textcolor{red}{18}\% &  & .230 &  & \textcolor{red}{20}\% &  & .220 &  & \textcolor{red}{8\%} &  & .261 &  & \textcolor{red}{9\%} &  & .164 &  & \textcolor{blue}{4\%} &  & .200 &  & \textcolor{blue}{4\%} &  & .130 &  & \textcolor{red}{43\%} &  & .156 &  & \textcolor{red}{41\%} &  & .277 &  & \textcolor{red}{9\%} &  & .312 &  & \textcolor{red}{10\%} \\
+AIGC-Synth. &  & .646 &  & \textcolor{red}{2\%} &  & .685 &  & \textcolor{red}{3\%} &  & .136 &  & \textcolor{red}{80\%} &  & .147 &  & \textcolor{red}{95\%} &  & .170 &  & \textcolor{red}{41\%} &  & .242 &  & \textcolor{red}{19\%} &  & .130 &  & \textcolor{red}{21\%} &  & .165 &  & \textcolor{red}{16\%} &  & .088 &  & \textcolor{red}{160\%} &  & .117 &  & \textcolor{red}{125\%} &  & .234 &  & \textcolor{red}{31\%} &  & .271 &  & \textcolor{red}{28\%} \\
+WSCL~\cite{wscl} &  & .668 &  & \textcolor{blue}{1\%} &  & .710 &  & \textcolor{blue}{1\%} &  & .240 &  & \textcolor{red}{2\%} &  & .281 &  & \textcolor{red}{2\%} &  & .245 &  & \textcolor{blue}{3\%} &  & .292 &  & \textcolor{blue}{2\%} &  & .204 &  & \textcolor{blue}{30\%} &  & .246 &  & \textcolor{blue}{29\%} &  & .238 &  & \textcolor{blue}{4\%} &  & .273 &  & \textcolor{blue}{4\%} &  & .319 &  & \textcolor{blue}{4\%} &  & .361 &  & \textcolor{blue}{4\%} \\
+MIMLv1 (\textit{Conf.}) &  & .691 &  & \textcolor{blue}{5\%} &  & .728 &  & \textcolor{blue}{4\%} &  & .302 &  & \textcolor{blue}{23\%} &  & .389 &  & \textcolor{blue}{36\%} &  & .353 &  & \textcolor{blue}{48\%} &  & .422 &  & \textcolor{blue}{47\%} &  & .547 &  & \textcolor{blue}{248\%} &  & .629 &  & \textcolor{blue}{228\%} &  & .397 &  & \textcolor{blue}{73\%} &  & .477 &  & \textcolor{blue}{81\%} &  & .458 &  & \textcolor{blue}{50\%} &  & .529 &  & \textcolor{blue}{53\%} \\
\rowcolor{gray!15} +MIMLv2 &  & .684 &  & \textcolor{blue}{4\%} &  & .720 &  & \textcolor{blue}{2\%} &  & \textbf{.368} &  & \textbf{\textcolor{blue}{50\%}} &  & \textbf{.443} &  & \textbf{\textcolor{blue}{55\%}} &  & .338 &  & \textcolor{blue}{41\%} &  & .416 &  & \textcolor{blue}{45\%} &  & .562 &  & \textcolor{blue}{258\%} &  & .644 &  & \textcolor{blue}{235\%} &  & .470 &  & \textcolor{blue}{105\%} &  & .548 &  & \textcolor{blue}{108\%} &  & .484 &  & \textcolor{blue}{58\%} &  & .554 &  & \textcolor{blue}{60\%} \\
\rowcolor{gray!15} +MIMLv2+O.J &  & \textbf{.703} &  & \textbf{\textcolor{blue}{7\%}} &  & \textbf{.742} &  & \textbf{\textcolor{blue}{6\%}} &  & .355 &  & \textcolor{blue}{45\%} &  & .437 &  & \textcolor{blue}{53\%} &  & \textbf{.358} &  & \textbf{\textcolor{blue}{50\%}} &  & \textbf{.440} &  & \textbf{\textcolor{blue}{53\%}} &  & \textbf{.575} &  & \textbf{\textcolor{blue}{266\%}} &  & \textbf{.656} &  & \textbf{\textcolor{blue}{242\%}} &  & \textbf{.497} &  & \textbf{\textcolor{blue}{117\%}} &  & \textbf{.575} &  & \textbf{\textcolor{blue}{119\%}} &  & \textbf{.498} &  & \textbf{\textcolor{blue}{63\%}} & \textbf{} & \textbf{.570} & \textbf{} & \textbf{\textcolor{blue}{65\%}} \\ \hline
TruFor B.L. &  & .755 &  & 0 &  & .794 &  & 0 &  & .363 &  & 0 &  & .411 &  & 0 &  & .320 &  & 0 &  & .377 &  & 0 &  & .248 &  & 0 &  & .289 &  & 0 &  & .296 &  & 0 &  & .346 &  & 0 &  & .396 &  & 0 &  & .443 &  & 0 \\
+PSCC-Synth. &  & .750 &  & \textcolor{red}{1\%} &  & .798 &  & \textcolor{red}{1\%} &  & .340 &  & \textcolor{red}{7\%} &  & .384 &  & \textcolor{red}{7\%} &  & .325 &  & \textcolor{blue}{2\%} &  & .373 &  & \textcolor{red}{1\%} &  & .253 &  & \textcolor{blue}{2\%} &  & .298 &  & \textcolor{blue}{3\%} &  & .278 &  & \textcolor{red}{6\%} &  & .331 &  & \textcolor{red}{5\%} &  & .389 &  & \textcolor{red}{2\%} &  & .437 &  & \textcolor{red}{2\%} \\
+DEFACTO\cite{defacto} &  & .752 &  & \textcolor{red}{0\%} &  & .791 &  & \textcolor{red}{0\%} &  & .322 &  & \textcolor{red}{13\%} &  & .357 &  & \textcolor{red}{15\%} &  & .214 &  & \textcolor{red}{50\%} &  & .260 &  & \textcolor{red}{45\%} &  & .209 &  & \textcolor{red}{19\%} &  & .247 &  & \textcolor{red}{17\%} &  & .184 &  & \textcolor{red}{61\%} &  & .217 &  & \textcolor{red}{59\%} &  & .336 &  & \textcolor{red}{18\%} &  & .374 &  & \textcolor{red}{18\%} \\
+AIGC-Synth. &  & .746 &  & \textcolor{red}{1\%} &  & .791 &  & \textcolor{red}{0\%} &  & .220 &  & \textcolor{red}{65\%} &  & .261 &  & \textcolor{red}{57\%} &  & .213 &  & \textcolor{red}{50\%} &  & .263 &  & \textcolor{red}{43\%} &  & .221 &  & \textcolor{red}{12\%} &  & .267 &  & \textcolor{red}{8\%} &  & .121 &  & \textcolor{red}{146\%} &  & .156 &  & \textcolor{red}{122\%} &  & .304 &  & \textcolor{red}{30\%} &  & .348 &  & \textcolor{red}{28\%} \\
+WSCL~\cite{wscl} &  & .758 &  & \textcolor{blue}{1\%} &  & .797 &  & \textcolor{blue}{1\%} &  & .362 &  & \textcolor{red}{0\%} &  & .410 &  & \textcolor{red}{0\%} &  & .326 &  & \textcolor{blue}{1\%} &  & .373 &  & \textcolor{blue}{1\%} &  & .307 &  & \textcolor{blue}{24\%} &  & .352 &  & \textcolor{blue}{22\%} &  & .310 &  & \textcolor{blue}{5\%} &  & .359 &  & \textcolor{blue}{4\%} &  & .412 &  & \textcolor{blue}{4\%} &  & .458 &  & \textcolor{blue}{3\%} \\
+MIMLv1 (\textit{Conf.}) &  & .755 &  & \textcolor{blue}{0\%} &  & .794 &  & \textcolor{blue}{0\%} &  & .385 &  & \textcolor{blue}{6\%} &  & .458 &  & \textcolor{blue}{10\%}&  & .352 &  & \textcolor{blue}{9\%} &  & .408 &  & \textcolor{blue}{8\%} &  & .596 &  & \textcolor{blue}{158\%}&  & .678 &  & \textcolor{blue}{157\%}&  & .340 &  & \textcolor{blue}{13\%}&  & .401 &  & \textcolor{blue}{14\%}&  & .486 &  & \textcolor{blue}{18\%}&  & .548 &  & \textcolor{blue}{19\%}\\
\rowcolor{gray!15} +MIMLv2 &  & .767&  & \textcolor{blue}{2\%} &  & .794 &  & \textcolor{blue}{0\%} &  & .387 &  & \textcolor{blue}{6\%} &  & .458 &  & \textcolor{blue}{10\%}&  & .407 &  & \textcolor{blue}{21\%}&  & .472 &  & \textcolor{blue}{20\%}&  & .618 &  & \textcolor{blue}{160\%}&  & .696 &  & \textcolor{blue}{158\%} &  & \textbf{.370} & \textbf{} & \textbf{\textcolor{blue}{20\%}} & \textbf{} & \textbf{.422} & \textbf{} & \textbf{\textcolor{blue}{18\%}} &  & .510 &  & \textcolor{blue}{22\%}&  & .568 &  & \textcolor{blue}{22\%}\\
\rowcolor{gray!15} +MIMLv2+O.J &  & \textbf{.770} & \textbf{} & \textbf{\textcolor{blue}{2\%}} & \textbf{} & \textbf{.811} & \textbf{} & \textbf{\textcolor{blue}{2\%}} &  & \textbf{.466} & \textbf{} & \textbf{\textcolor{blue}{22\%}} & \textbf{} & \textbf{.511} & \textbf{} & \textbf{\textcolor{blue}{20\%}} & \textbf{} & \textbf{.439} & \textbf{} & \textbf{\textcolor{blue}{27\%}} & \textbf{} & \textbf{.502} & \textbf{} & \textbf{\textcolor{blue}{25\%}} & \textbf{} & \textbf{.619} & \textbf{} & \textbf{\textcolor{blue}{160\%}} & \textbf{} & \textbf{.700} & \textbf{} & \textbf{\textcolor{blue}{159\%}} &  & .309 &  & \textcolor{blue}{4\%} &  & .357 &  & \textcolor{blue}{3\%} &  & \textbf{.521} & \textbf{} & \textbf{\textcolor{blue}{24\%}} & \textbf{} & \textbf{.576} & \textbf{} & \textbf{\textcolor{blue}{23\%}} \\ \hline
SparseViT B.L. &  & .702 &  & 0 &  & .747 &  & 0 &  & .320 &  & 0 &  & .352 &  & 0 &  & .288 &  & 0 &  & .343 &  & 0 &  & .199 &  & 0 &  & .239 &  & 0 &  & .177 &  & 0 &  & .232 &  & 0 &  & .337 &  & 0 &  & .383 &  & 0 \\
+PSCC-Synth. &  & .627 &  & \textcolor{red}{11\%} &  & .677 &  & \textcolor{red}{9\%} &  & .304 &  & \textcolor{red}{5\%} &  & .333 &  & \textcolor{red}{5\%} &  & .256 &  & \textcolor{red}{11\%} &  & .311 &  & \textcolor{red}{9\%} &  & .196 &  & \textcolor{red}{1\%} &  & .243 &  & \textcolor{blue}{1\%} &  & .169 &  & \textcolor{red}{5\%} &  & .232 &  & \textcolor{red}{0\%} &  & .310 &  & \textcolor{red}{8\%} &  & .359 &  & \textcolor{red}{6\%} \\
+DEFACTO\cite{defacto} &  & .692 &  & \textcolor{red}{1\%} &  & .730 &  & \textcolor{red}{2\%} &  & .257 &  & \textcolor{red}{20\%} &  & .283 &  & \textcolor{red}{20\%} &  & .310 &  & \textcolor{blue}{8\%} &  & .364 &  & \textcolor{blue}{6\%} &  & .186 &  & \textcolor{red}{7\%} &  & .225 &  & \textcolor{red}{6\%} &  & .155 &  & \textcolor{red}{12\%} &  & .198 &  & \textcolor{red}{15\%} &  & .320 &  & \textcolor{red}{5\%} &  & .362 &  & \textcolor{red}{6\%} \\
+AIGC-Synth. &  & .533 &  & \textcolor{red}{24\%} &  & .583 &  & \textcolor{red}{22\%} &  & .119 &  & \textcolor{red}{63\%} &  & .141 &  & \textcolor{red}{60\%} &  & .133 &  & \textcolor{red}{54\%} &  & .174 &  & \textcolor{red}{49\%} &  & .088 &  & \textcolor{red}{56\%} &  & .110 &  & \textcolor{red}{54\%} &  & .033 &  & \textcolor{red}{81\%} &  & .045 &  & \textcolor{red}{81\%} &  & .181 &  & \textcolor{red}{46\%} &  & .211 &  & \textcolor{red}{45\%} \\
+WSCL~\cite{wscl} &  & .709 &  & \textcolor{blue}{1\%} &  & .755 &  & \textcolor{blue}{1\%} &  & .324 &  & \textcolor{blue}{1\%} &  & .356 &  & \textcolor{blue}{1\%} &  & .301 &  & \textcolor{blue}{4\%} &  & .352 &  & \textcolor{blue}{3\%} &  & .267 &  & \textcolor{blue}{34\%} &  & .314 &  & \textcolor{blue}{31\%} &  & .184 &  & \textcolor{blue}{4\%} &  & .241 &  & \textcolor{blue}{4\%} &  & .357 &  & \textcolor{blue}{6\%} &  & .404 &  & \textcolor{blue}{6\%} \\
+MIMLv1 (\textit{Conf.}) &  & .714 &  & \textcolor{blue}{2\%} &  & .762 &  & \textcolor{blue}{2\%} &  & .394 &  & \textcolor{blue}{23\%} &  & .458 &  & \textcolor{blue}{30\%}&  & .306 &  & \textcolor{blue}{6\%} &  & .359 &  & \textcolor{blue}{5\%} &  & .535 &  & \textcolor{blue}{169\%}&  & .617 &  & \textcolor{blue}{158\%}&  & .215 &  & \textcolor{blue}{21\%}&  & .270 &  & \textcolor{blue}{16\%}&  & .433 &  & \textcolor{blue}{28\%}&  & .493 &  & \textcolor{blue}{29\%}\\
\rowcolor{gray!15} +MIMLv2 &  & \textbf{.723} &  & \textbf{\textcolor{blue}{3\%}} &  & \textbf{.771} &  & \textbf{\textcolor{blue}{3\%}} &  & .441 &  & \textcolor{blue}{38\%} &  & .509 &  & \textcolor{blue}{45\%}&  & .338 &  & \textcolor{blue}{17\%}&  & .396 &  & \textcolor{blue}{15\%}&  & .544 &  & \textcolor{blue}{173\%}&  & .624 &  & \textcolor{blue}{161\%} &  & \textbf{.234} & \textbf{} & \textbf{\textcolor{blue}{32\%}} & \textbf{} & \textbf{.289} & \textbf{} & \textbf{\textcolor{blue}{25\%}} &  & .456 &  & \textcolor{blue}{35\%}&  & .518 &  & \textcolor{blue}{35\%}\\
\rowcolor{gray!15} +MIMLv2+O.J &  & .720 & \textbf{} & \textcolor{blue}{3\%} & \textbf{} & \textcolor{blue}{.767} & \textbf{} & \textcolor{blue}{3\%} &  & \textbf{.477} & \textbf{} & \textbf{\textcolor{blue}{49\%}} & \textbf{} & \textbf{.547} & \textbf{} & \textbf{\textcolor{blue}{55\%}} & \textbf{} & \textbf{.346} & \textbf{} & \textbf{\textcolor{blue}{20\%}} & \textbf{} & \textbf{.409} & \textbf{} & \textbf{\textcolor{blue}{19\%}} & \textbf{} & \textbf{.551} & \textbf{} & \textbf{\textcolor{blue}{178\%}} & \textbf{} & \textbf{{.633}} & \textbf{} & \textbf{\textcolor{blue}{165\%}} &  & \textbf{.252} &  & \textbf{\textcolor{blue}{42\%}} &  & \textbf{.325} &  & \textcolor{blue}{\textbf{40\%}} &  & \textbf{.469} & \textbf{} & \textbf{\textcolor{blue}{40\%}} & \textbf{} & \textbf{.536} & \textbf{} & \textbf{\textcolor{blue}{40\%}} \\ \hline
IMDPrompter B.L. &  & .718 &  & 0 &  & .765 &  & 0 &  & .463 &  & 0 &  & .522 &  & 0 &  & .346 &  & 0 &  & .419 &  & 0 &  & .261 &  & 0 &  & .310 &  & 0 &  & .354 &  & 0 &  & .414 &  & 0 &  & .428 &  & 0 &  & .486 &  & 0 \\
+PSCC-Synth. &  & .682 &  & \textcolor{red}{5\%} &  & .729 &  & \textcolor{red}{5\%} &  & .473 &  & \textcolor{blue}{2\%} &  & .531 &  & \textcolor{blue}{2\%} &  & .295 &  & \textcolor{red}{15\%} &  & .356 &  & \textcolor{red}{15\%} &  & .262 &  & \textcolor{blue}{0\%} &  & .312 &  & \textcolor{blue}{1\%} &  & .286 &  & \textcolor{red}{19\%} &  & .337 &  & \textcolor{red}{19\%} &  & .400 &  & \textcolor{red}{7\%} &  & .453 &  & \textcolor{red}{7\%} \\
+DEFACTO\cite{defacto} &  & .701 &  & \textcolor{red}{2\%} &  & .749 &  & \textcolor{red}{2\%} &  & .439 &  & \textcolor{red}{5\%} &  & .508 &  & \textcolor{red}{3\%} &  & .284 &  & \textcolor{red}{18\%} &  & .347 &  & \textcolor{red}{17\%} &  & .243 &  & \textcolor{red}{7\%} &  & .286 &  & \textcolor{red}{8\%} &  & .281 &  & \textcolor{red}{21\%} &  & .329 &  & \textcolor{red}{21\%} &  & .390 &  & \textcolor{red}{9\%} &  & .444 &  & \textcolor{red}{9\%} \\
+AIGC-Synth. &  & .630 &  & \textcolor{red}{12\%} &  & .678 &  & \textcolor{red}{11\%} &  & .254 &  & \textcolor{red}{45\%} &  & .289 &  & \textcolor{red}{45\%} &  & .218 &  & \textcolor{red}{37\%} &  & .263 &  & \textcolor{red}{37\%} &  & .222 &  & \textcolor{red}{15\%} &  & .264 &  & \textcolor{red}{15\%} &  & .263 &  & \textcolor{red}{26\%} &  & .309 &  & \textcolor{red}{25\%} &  & .317 &  & \textcolor{red}{26\%} &  & .361 &  & \textcolor{red}{26\%} \\
+WSCL~\cite{wscl} &  & .701 &  & \textcolor{red}{2\%} &  & .750 &  & \textcolor{red}{2\%} &  & .485 &  & \textcolor{blue}{5\%} &  & .543 &  & \textcolor{blue}{4\%} &  & .360 &  & \textcolor{blue}{4\%} &  & .433 &  & \textcolor{blue}{3\%} &  & .285 &  & \textcolor{blue}{9\%} &  & .336 &  & \textcolor{blue}{8\%} &  & .356 &  & \textcolor{blue}{1\%} &  & .417 &  & \textcolor{blue}{1\%} &  & .437 &  & \textcolor{blue}{2\%} &  & .496 &  & \textcolor{blue}{2\%} \\
+MIMLv1 (\textit{Conf.}) &  & .724 &  & \textcolor{blue}{1\%} &  & .769 &  & \textcolor{blue}{1\%} &  & .542 &  & \textcolor{blue}{17\%} &  & .610 &  & \textcolor{blue}{17\%}&  & .418 &  & \textcolor{blue}{21\%} &  & .501 &  & \textcolor{blue}{20\%} &  & .601 &  & \textcolor{blue}{130\%}&  & .696 &  & \textcolor{blue}{125\%}&  & .392 &  & \textcolor{blue}{11\%}&  & .456 &  & \textcolor{blue}{10\%}&  & .535 &  & \textcolor{blue}{25\%}&  & .606 &  & \textcolor{blue}{25\%}\\
\rowcolor{gray!15} +MIMLv2 &  & \textbf{.729} &  & \textbf{\textcolor{blue}{2\%}} &  & \textbf{.773} &  & \textbf{\textcolor{blue}{1\%}} &  & .564 &  & \textcolor{blue}{22\%} &  & .635 &  & \textcolor{blue}{22\%}&  & .426 &  & \textcolor{blue}{23\%}&  & .512 &  & \textcolor{blue}{22\%}&  & .614 &  & \textcolor{blue}{135\%}&  & .708 &  & \textcolor{blue}{128\%} &  & \textbf{.418} & \textbf{} & \textbf{\textcolor{blue}{18\%}} & \textbf{} & \textbf{.490} & \textbf{} & \textbf{\textcolor{blue}{18\%}} &  & .550 &  & \textcolor{blue}{29\%}&  & .624 &  & \textcolor{blue}{28\%}\\
\rowcolor{gray!15} +MIMLv2+O.J &  & .735 & \textbf{} & \textcolor{blue}{2\%} & \textbf{} & \textcolor{blue}{.778} & \textbf{} & \textcolor{blue}{2\%} &  & \textbf{.626} & \textbf{} & \textbf{\textcolor{blue}{35\%}} & \textbf{} & \textbf{.706} & \textbf{} & \textbf{\textcolor{blue}{35\%}} & \textbf{} & \textbf{.433} & \textbf{} & \textbf{\textcolor{blue}{25\%}} & \textbf{} & \textbf{.521} & \textbf{} & \textbf{\textcolor{blue}{24\%}} & \textbf{} & \textbf{.635} & \textbf{} & \textbf{\textcolor{blue}{143\%}} & \textbf{} & \textbf{{.721}} & \textbf{} & \textbf{\textcolor{blue}{133\%}} &  & \textbf{.425} &  & \textbf{\textcolor{blue}{20\%}} &  & \textbf{.498} &  & \textcolor{blue}{\textbf{20\%}} &  & \textbf{.571} & \textbf{} & \textbf{\textcolor{blue}{33\%}} & \textbf{} & \textbf{.645} & \textbf{} & \textbf{\textcolor{blue}{33\%}} \\ \hline
\end{tabular}
\label{tab: othermodedataabl}
\end{table*}

\section{Experiments}~\label{sec. 7}
\subsection{Experiments on the Web-IML Model (IML Task)}
This subsection evaluates the effectiveness of the proposed Web-IML model by comparing it with other IML models on multiple real-world handcrafted and AIGC forgery benchmarks. We also perform an ablation study to provide an in-depth analysis of the Web-IML model.

\smallskip

\noindent\textbf{Implementation Details}. For Web-IML, we adopt ConvNeXt-Base~\cite{liu2022convnet} as the feature extractor. The model is trained for 200k iterations with a batch size of 16, using a 512$\times$512 input size, consistent with previous works~\cite{sparsevit,guillaro2023trufor}. We employ cross entropy loss and AdamW~\cite{adamw} optimizer, with the learning rate linearly decaying from 1e-4 to 1e-6. CASIAv2~\cite{casia} and tampCOCO~\cite{catnet} serve as baseline training sets. For fair comparison, mask predictions are binarized with a fixed threshold of 0.5 and evaluated using IoU and binary F1-score.

\vspace{+0.1cm}
\smallskip

\begin{figure}[t]
 	\centering
 	\includegraphics[width=1.0\linewidth]{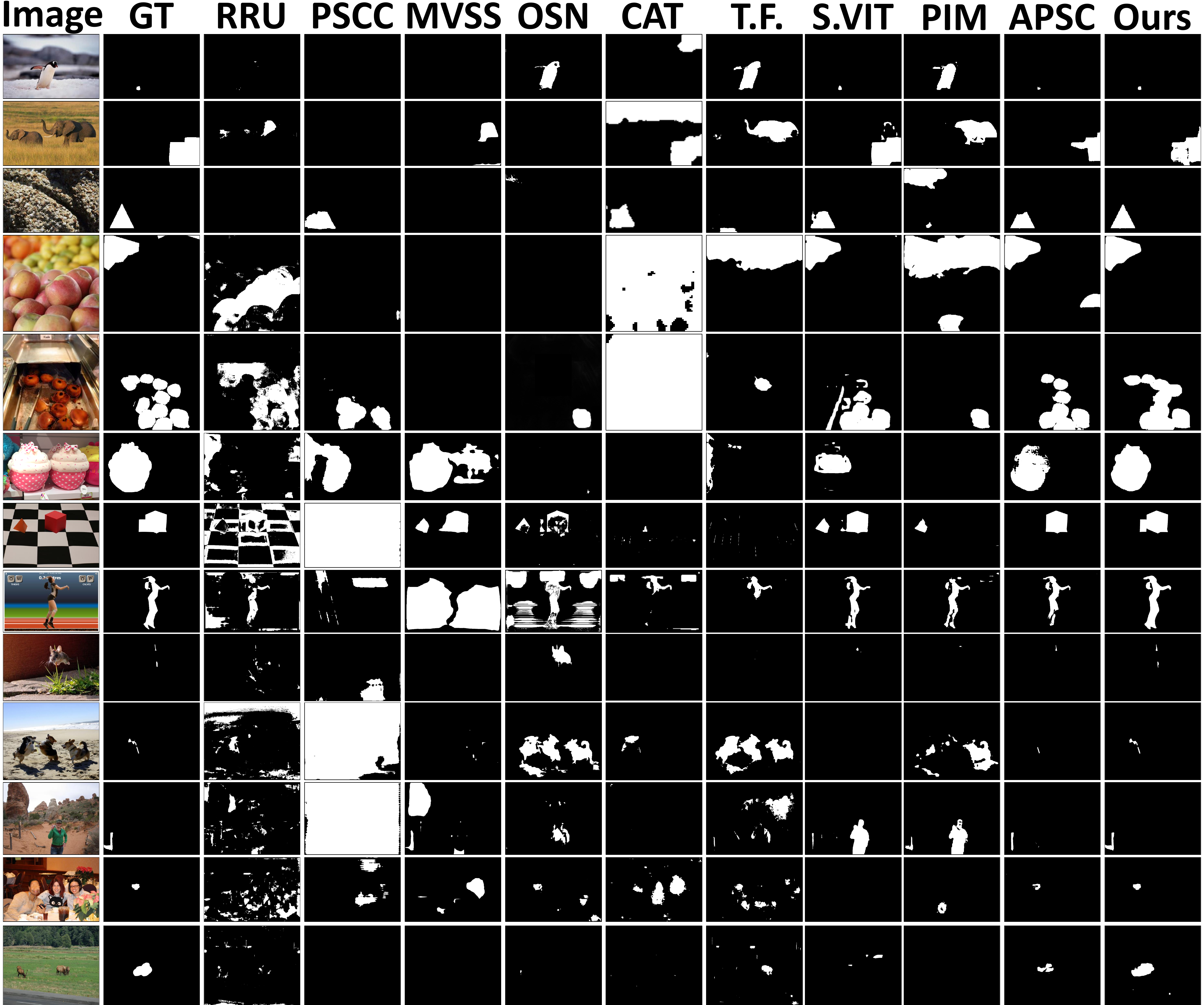}
 	\caption{Qualitative evaluation on image manipulation localization. From left to right: input image, Ground-Truth (abbreviated as GT), RRU-Net (RRU), PSCC-Net (PSCC), MVSS-Net (MVSS), IF-OSN (OSN), CAT-Net (CAT), TruFor (T.F.), SparseViT (S.ViT), APSC-Net (APSC) and ours.
 	}
 \label{fig:Fig12_VIZ}
 \end{figure}

\noindent\textbf{Comparison Study}. Table~\ref{tab: maincomp} shows that our Web-IML significantly outperforms existing methods. For example, it achieves an average IoU that is 21.6 points higher than SparseViT. \textit{Furthermore, Web-IML consistently outperforms our conference version, APSC-Net, across various settings}. These results demonstrate the strong generalization capabilities of our Web-IML model. Qualitative comparison is illustrated in Fig.~\ref{fig:Fig12_VIZ}.

\vspace{+0.1cm}
\smallskip

\noindent\textbf{Ablation Study}. The ablation results are shown in Table~\ref{tab: modelabl}. Setting (2), which includes the proposed Multi-Scale Perception (MP) module, achieves 5.2 points higher IoU than the baseline (1). This indicates that the proposed MP module effectively integrates high-level and low-level features to detect diverse image artifacts. Setting (3), which further incorporates the Self-Rectification (SR) module, yields an additional 3.9 points IoU improvement over Setting (2), as SR enables the model to detect and correct its errors. Setting (4) outperforms (3) via our nested attention module that enables in-depth analysis. Setting (5) outperforms (4) by including a second round of self-rectification refinement, demonstrating that further refinement can minimize incorrect predictions.

\vspace{+0.1cm}
\smallskip

\noindent\textbf{Robustness Evaluation}. We evaluate Web-IML under various distortions, including image resizing (scale factors 0.78 and 0.25), Gaussian blur (kernel sizes 3 and 15), and JPEG compression (quality factors 50 and 100). The results are shown in Table~\ref{tab: robust}. Our model exhibits stable performance under these conditions, demonstrating strong robustness.

\subsection{Experiments on the Web-Supervision Method (IML Task)}
This subsection evaluates the effectiveness of our proposed web-supervision approach in addressing the data scarcity issue for IML. Our web-supervision encompasses both the MIMLv2 dataset and the Object Jitter method. We compare the performance of the same IML model trained on data created by our web-supervision approach against other synthetic methods and WSCL~\cite{wscl}, a representative weakly supervised framework.

\smallskip
\vspace{+0.1cm}

\noindent \textbf{Implementation Details}. The implementation details remain the same as in the previous subsection VII.A. When supplementing the baseline training sets with additional IML training sets, the total training volume is kept constant with a uniform sampling ratio across all datasets. For models integrated with WSCL, training involved two types of data: direct training with the fully (pixel-level) annotated tampCOCO~\cite{catnet} and CASIAv2~\cite{casia} datasets, and WSCL-specific training with weakly (image-level) annotated MIMLv2 and SA~\cite{SAM} dataset images. The training volume and other configurations remain the same as ours.

\smallskip
\vspace{+0.1cm}

\noindent \textbf{Comparison Study for the MIMLv2 Dataset}. Table~\ref{tab: dataabl} presents the IML performance of the Web-IML model when trained on different datasets. The baseline (1) uses the two most common training sets: tampCOCO~\cite{catnet} and CASIAv2~\cite{casia}. Incorporating synthetic training data from PSCC-Net~\cite{liu2022pscc}, DEFACTO~\cite{defacto} or OpenSDID~\cite{wang2025opensdi} diminishes the average model performance (settings 2, 3, 4). This is because the first two synthetic datasets do not notably surpass the baseline data in quality or diversity, while the AIGC forgeries from OpenSDID~\cite{wang2025opensdi} differ significantly from non-AIGC forgeries. These results confirm that simply increasing the scale of training data does not lead to improved performance. In contrast, including the MIMLv1 (setting 6) or MIMLv2 (setting 7) datasets significantly improves model performance by an average of 10 and 12 points, respectively, without increasing training or inference burden. Additionally, although WSCL (setting 5) offers improvements over baselines, its effectiveness is considerably lower than that of our web-supervision. This is because our method explicitly produces accurate mask annotations, by leveraging the image correspondence that substantially reduces task difficulty. Conversely, WSCL lacks the ability to leverage this valuable information, and relies heavily on unstable noise domain features. These results confirm that our MIMLv1 and MIMLv2 datasets considerably alleviate data scarcity and overfitting issues in real-world scenarios. This is attributable to their large scale, manual forgery, and considerably higher quality and diversity compared to other datasets. \textit{Notably, MIMLv2 yields greater improvement due to its larger quantity and superior diversity compared to MIMLv1.}

\smallskip
\vspace{+0.1cm}

\noindent \textbf{Comparison Study for the Object Jitter}. Table~\ref{tab: dataabl} shows that including the training data generated by our Object Jitter method further improves the model when combined with MIMLv1 (setting 8 outperforms 6) or MIMLv2 (setting 10 outperforms 7) datasets. This improvement is attributed to Object Jitter's consistent generation of high-quality hard samples, which effectively supplements the training data by compensating for the relative scarcity of copy-move forgeries.

\smallskip
\vspace{+0.1cm}

\noindent \textbf{Ablation Study for the Object Jitter}. Table~\ref{tab: ojabl} presents the ablation study for Object Jitter operations and image source. Setting (1) is the baseline, where Web-IML is trained with MIMLv2 but without Object Jitter. Setting (7) represents the full Object Jitter method. Settings (2, 3, 4) involve removing the size jitter, texture jitter, and exposure jitter operations, respectively; all of them result in worse performance than setting (7). 
These results demonstrate that all three operations contribute to improved performance through increased data diversity. Notably, Setting (2), which lacks size jitter, is significantly worse than (7). Meanwhile, Setting (5), which uses only size jitter, performs close to (7). This indicates that the size jitter operation is the primary contributor to the improvement. Since most Internet forgeries are not generated by copy-move, they exhibit exposure and texture artifacts. The size jitter operation, without altering the object's exposure or texture, reduces the model's reliance on exposure or texture cues, thus better supplementing MIMLv2. Furthermore, Setting (6) applies the full Object Jitter method using COCO~\cite{lin2014microsoft} as the image source, and setting (7) uses 400k random images from SA~\cite{SAM}. The similar performance of Settings (6) and (7) demonstrates that the Object Jitter method is robust to the choice of image source.

\vspace{+0.1cm}
\smallskip

\noindent\textbf{Verification of Conclusions on Other IML Models}. We extended our verification process to other IML models. Specifically, we evaluated MIMLv2 and Object Jitter on PSCC-Net~\cite{liu2022pscc}, CAT-Net~\cite{catnet}, TruFor~\cite{guillaro2023trufor}, SparseViT~\cite{sparsevit} and IMDPrompter~\cite{zhangimdprompter} (a large model based on SAM~\cite{SAM}) using the same training data and configuration as ours. The results in Table~\ref{tab: othermodedataabl} clearly indicate that MIMLv1, MIMLv2, and Object Jitter all lead to significantly higher scores. \textit{MIMLv2, in particular, consistently yields significantly greater improvement than MIMLv1 (conference version)}. Moreover, \textit{Object Jitter further boosts performance when integrated with MIMLv2}. These results robustly confirm our previous analyses and conclusions.

\begin{table}[]
\setlength{\tabcolsep}{0.8pt}
\caption{Comparison study of CIML on SPG (left) and SDG (right). '*' denotes model trained with both SDG and SPG data. 'Nonzero' denotes using the non-zero region of the difference map as mask. 'QES' denotes the predictions filtered by our QES metric.}
\begin{tabular}{ccccccccccccccccccc}
\cline{1-9} \cline{11-19}
\multicolumn{9}{c}{CIML on SPG} & & \multicolumn{9}{c}{CIML on SDG} \\
\cline{1-9} \cline{11-19}
Method &  & P &  & R &  & F &  & IoU &  & Method &  & P &  & R &  & F &  & IoU \\ \cline{1-1} \cline{3-3} \cline{5-5} \cline{7-7} \cline{9-9} \cline{11-11} \cline{13-13} \cline{15-15} \cline{17-17} \cline{19-19} 
Nonzero &  & .159 &  & .987 &  & .273 &  & .079 &  & DMVN*~\cite{dmvn} &  & .553 &  & .357 &  & .434 &  & .276 \\
OTSU~\cite{otsu} &  & .812 &  & .589 &  & .683 &  & .497 &  & DMVN~\cite{dmvn} &  & .438 &  & .568 &  & .495 &  & .317 \\
DMVN*~\cite{dmvn} &  & .329 &  & .618 &  & .430 &  & .276 &  & DMAC*~\cite{dmac} &  & .583 &  & .538 &  & .559 &  & .410 \\
DMVN~\cite{dmvn} &  & .682 &  & .781 &  & .728 &  & .578 &  & DMAC~\cite{dmac} &  & .703 &  & .622 &  & .660 &  & .518 \\
\cline{11-19}
DMAC*~\cite{dmac} &  & .617 &  & .624 &  & .620 &  & .432 &  & SACM~\cite{mimlv1} (\textit{Conf.}) &  & .778 &  & .819 &  & .798 &  & .702 \\
DMAC~\cite{dmac} &  & .746 &  & .710 &  & .728 &  & .573 &  & \cellcolor{gray!15}{Corr-DINO (Ours)} & \cellcolor{gray!15}{} & \cellcolor{gray!15}{.816} & \cellcolor{gray!15}{} & \cellcolor{gray!15}{.862} & \cellcolor{gray!15}{} & \cellcolor{gray!15}{.838} & \cellcolor{gray!15}{} & \cellcolor{gray!15}{.744} \\
DASS (Ours) &  & \textbf{.860} & \textbf{} & \textbf{.921} & \textbf{} & \textbf{.889} & \textbf{} & \textbf{.835} &  & \cellcolor{gray!15}{Corr-DINO (QES)} & \cellcolor{gray!15}{} & \cellcolor{gray!15}{\textbf{.939}} & \cellcolor{gray!15}{} & \cellcolor{gray!15}{\textbf{.971}} & \cellcolor{gray!15}{} & \cellcolor{gray!15}{\textbf{.954}} & \cellcolor{gray!15}{} & \cellcolor{gray!15}{\textbf{.912}} \\ \cline{1-9} \cline{11-19} 
\end{tabular}
\label{tab: spgcomp}
\end{table}

\begin{table}[]
\setlength{\tabcolsep}{2.5pt}
\caption{Ablation study of CIML on SPG. 'Forged' denotes using the forged image as input, 'Auth.' denotes using the authentic image as input, 'Diff.' denotes using the difference map between the forged and authentic image as input.}
\begin{tabular}{ccccccccccccccccc}
\hline
Num. &  & Backbone &  & Forged &  & Auth. &  & Diff. &  & P &  & R &  & F1 &  & IoU \\ \cline{1-1} \cline{3-3} \cline{5-5} \cline{7-7} \cline{9-9} \cline{11-11} \cline{13-13} \cline{15-15} \cline{17-17} 
(1) &  & VGG~\cite{vgg} &  & \checkmark &  & $\times$ &  & $\times$ &  & .423 &  & .365 &  & .392 &  & .236 \\
(2) &  & VGG~\cite{vgg} &  & \checkmark &  & \checkmark &  & $\times$ &  & .800 &  & .857 &  & .827 &  & .718 \\
(3) &  & VGG~\cite{vgg} &  & $\times$ &  & $\times$ &  & \checkmark &  & .771 &  & .857 &  & .812 &  & .741 \\
(4) &  & VGG~\cite{vgg} &  & \checkmark &  & \checkmark &  & \checkmark &  & .825 &  & .899 &  & .860 &  & .781 \\ \hline
(5) &  & VAN~\cite{van} &  & \checkmark &  & $\times$ &  & $\times$ &  & .573 &  & .533 &  & .552 &  & .408 \\
(6) &  & VAN~\cite{van} &  & \checkmark &  & \checkmark &  & \checkmark &  & \textbf{.860} & \textbf{} & \textbf{.921} & \textbf{} & \textbf{.889} & \textbf{} & \textbf{.835} \\ \hline
\end{tabular}
\label{tab: spgabl}
\end{table}

\subsection{Experiments on the CAAAv2 Models (CIML task)}
This subsection evaluates the effectiveness of the proposed CAAAv2 in the CIML task by comparing its DASS and Corr-DINO models against other CIML models. A higher CIML score means more accurate auto-annotation. We also conduct detailed ablation studies for in-depth analysis.

\vspace{+0.1cm}
\smallskip

\noindent \textbf{Implementation Details}. Given that real-world forgeries in the IMD20~\cite{imd20} dataset closely resemble the target images for annotation, we use a portion of this dataset to evaluate model performance using IoU and binary F1-score metrics. We categorize the forged images in IMD20 into SPG or SDG and randomly split them into training and test sets with a 3:1 ratio. CASIAv2~\cite{casia} and synthetic COCO~\cite{lin2014microsoft} datasets are also used for training. Input images are resized to 512$\times$512, and a consistent training pipeline is applied across all methods.

\vspace{+0.1cm}
\smallskip

\noindent\textbf{Comparison Study}. Table~\ref{tab: spgcomp} presents comparison results for SPG (left) and SDG (right) separately. Our method clearly outperforms these previous methods because our category-aware and prior-feature-denoising paradigm significantly reduces task difficulty. As shown in the right side of Table~\ref{tab: spgcomp}, our Corr-DINO, prior to QES filtering, achieves an IoU of 0.744. \textit{This is 4.2 points higher than the SACM~\cite{mimlv1} (our conference version)}, demonstrating that our frozen-denoising paradigm notably alleviates the overfitting issue when training on limited data. Notably, DMVN~\cite{dmvn} and DMAC~\cite{dmac} models trained on both SPG and SDG data perform worse than those trained exclusively on one category. This is attributed to the confusion caused by similar backgrounds in SPGs, confirming the necessity of our category-aware approach.

\vspace{+0.1cm}
\smallskip

\noindent\textbf{Ablation Study for the DASS Model Input}. Table~\ref{tab: spgabl} presents the results, confirming that both the prior feature difference map and the image pair are essential for accurate prediction in SPG. The difference map between forged and authentic images can roughly indicate the forged regions, while the images themselves provide semantic information to help the model denoise the difference map.

\vspace{+0.1cm}
\smallskip

\noindent\textbf{Ablation Study for the Corr-DINO Model Prior Feature Extractor}. The results are shown in Table~\ref{tab: sdgabl}, settings (1-24). For ConvNeXt, the trainable version performs better. For all ViT-based models, however, frozen versions perform significantly better. Since ViT models tend to overfit with limited training data, freezing the backbone alleviates catastrophic forgetting and overfitting. LoRA~\cite{hu2022lora} fine-tuned versions exhibit similar performance to their frozen counterparts, as LoRA also mitigates forgetting. The frozen DINOv2-Base~\cite{oquab2023dinov2} backbone performs optimally and is therefore adopted.

\vspace{+0.1cm}
\smallskip

\noindent\textbf{Ablation Study of the Corr-DINO Model Modules}. The results are shown in the bottom of Table~\ref{tab: sdgabl}, setting (25) is the baseline without any proposed modules, setting (29) is the full model with all proposed modules. In settings (26, 27, 28), we remove the proposed Multi-Aspect Denoiser, Feature Super Resolution and Learnable Aggregation modules from the full model (29) respectively. Setting (29) significantly outperforms (26). This validates that correlation features extracted from the frozen DINOv2 are highly noisy, and that denoising through multi-aspect feature integration is essential for accurate predictions. Setting (29) also significantly outperforms (27). This confirms that initial correlation features are too coarse for precise mask predictions, and our Feature Super Resolution module effectively addresses this by reconstructing fine-grained details. Finally, Setting (29) is superior to (28) because our learnable aggregation mitigates information loss inherent in previous rule-based aggregation.

\vspace{+0.1cm}
\smallskip

\noindent\textbf{Ablation Study of the QES}. The last two rows of Table~\ref{tab: sdgabl} show the results. The results of setting (30) are obtained by evaluating only the predictions from setting (29) where QES$>$0.5. QES-filtered predictions from setting (30) achieve an IoU of 0.912, representing a 16.8 point increase over setting (29). This confirms that QES effectively filters out unsatisfactory predictions, thereby notably improving annotation quality. 
Table~\ref{tab: qesabl} confirms QES's high relevance to the IoU metric without relying on ground-truth for evaluation.

\vspace{+0.1cm}
\smallskip

\noindent \textbf{SPG/SDG Classification}. We evaluate the performance of the SPG/SDG classification model within the proposed CAAAv2 on IMD20~\cite{imd20} dataset. Considering that a very small proportion of SPG image pairs are not spatially aligned, which could negatively impact the prediction's quality, we also include an additional linear binary classification layer to filter them out. The classification results are presented in Table~\ref{tab: spgsdg_exp}. It is evident that the voting ensemble of the classification models we used is accurate enough for SPG/SDG classification, as analyzed in Section III.A. In practice, to further ensure accuracy, we only keep images with a classification confidence above 99\% and discard the rest. Our QES also effectively filters the annotations of potentially misclassified samples.

\smallskip

\noindent\textbf{Our Models are Adequate Auto-Annotators}. In Table~\ref{tab: spgcomp}, both our DASS and Corr-DINO achieve high IoU scores. Considering the errors in IMD20's ground-truth~\cite{ji2023uncertainty}, especially in SPG images, our actual accuracy is higher. Clearly, our methods are sufficient for obtaining accurate auto-annotations.

\begin{table}[t!]
\setlength{\tabcolsep}{1pt}
\caption{Ablation study of CIML on SDG. `Freeze` denotes freezing the backbone and `L` denotes LoRA~\cite{hu2022lora} fine-tuning. `MAD` denotes the Multi-Aspect Denoiser. `FSR` denotes the Feature Super Resolution. `LA` denotes the Learnable Aggregation. `QES` denotes filtering the prediction with QES.}
\begin{tabular}{ccccccccccccccccccccc}
\hline
Num. &  & Backbone &  & Freeze &  & MAD &  & FSR &  & LA &  & QES &  & P &  & R &  & F1 &  & IoU \\ \cline{1-1} \cline{3-3} \cline{5-5} \cline{7-7} \cline{9-9} \cline{11-11} \cline{13-13} \cline{15-15} \cline{17-17} \cline{19-19} \cline{21-21} 
(1) &  & ConvNeXt-B~\cite{liu2022convnet} &  & $\times$ &  & \checkmark &  & \checkmark &  & \checkmark &  & $\times$ &  & .740 &  & .690 &  & .714 &  & .595 \\
(2) &  & ConvNeXt-B~\cite{liu2022convnet} &  & \checkmark &  & \checkmark &  & \checkmark &  & \checkmark &  & $\times$ &  & .635 &  & .554 &  & .592 &  & .414 \\
(3) &  & Eff.SAM~\cite{effsam} &  & $\times$ &  & \checkmark &  & \checkmark &  & \checkmark &  & $\times$ &  & .483 &  & .624 &  & .544 &  & .381 \\
(4) &  & Eff.SAM~\cite{effsam} &  & \checkmark &  & \checkmark &  & \checkmark &  & \checkmark &  & $\times$ &  & .606 &  & .502 &  & .549 &  & .403 \\
(5) &  & CLIP-L~\cite{clip} &  & $\times$ &  & \checkmark &  & \checkmark &  & \checkmark &  & $\times$ &  & .665 &  & .509 &  & .577 &  & .453 \\
(6) &  & CLIP-L~\cite{clip} &  & L &  & \checkmark &  & \checkmark &  & \checkmark &  & $\times$ &  & .744 &  & .736 &  & .740 &  & .587 \\
(7) &  & CLIP-L~\cite{clip} &  & \checkmark &  & \checkmark &  & \checkmark &  & \checkmark &  & $\times$ &  & .749 &  & .721 &  & .735 &  & .596 \\
(8) &  & DINOv1-B~\cite{dinov1} &  & $\times$ &  & \checkmark &  & \checkmark &  & \checkmark &  & $\times$ &  & .693 &  & .512 &  & .589 &  & .456 \\
(9) &  & DINOv1-B~\cite{dinov1} &  & L &  & \checkmark &  & \checkmark &  & \checkmark &  & $\times$ &  & .693 &  & .742 &  & .717 &  & .595 \\
(10) &  & DINOv1-B~\cite{dinov1} &  & \checkmark &  & \checkmark &  & \checkmark &  & \checkmark &  & $\times$ &  & .700 &  & .700 &  & .700 &  & .556 \\
(11) &  & DINOv2-S~\cite{oquab2023dinov2} &  & $\times$ &  & \checkmark &  & \checkmark &  & \checkmark &  & $\times$ &  & .739 &  & .653 &  & .693 &  & .584 \\
(12) &  & DINOv2-S~\cite{oquab2023dinov2} &  & L &  & \checkmark &  & \checkmark &  & \checkmark &  & $\times$ &  & .783 &  & .733 &  & .758 &  & .637 \\
(13) &  & DINOv2-S~\cite{oquab2023dinov2} &  & \checkmark &  & \checkmark &  & \checkmark &  & \checkmark &  & $\times$ &  & .777 &  & .781 &  & .772 &  & .667 \\
(14) &  & DINOv2-B~\cite{oquab2023dinov2} &  & $\times$ &  & \checkmark &  & \checkmark &  & \checkmark &  & $\times$ &  & .705 &  & .574 &  & .633 &  & .472 \\
(15) &  & DINOv2-B~\cite{oquab2023dinov2} &  & L &  & \checkmark &  & \checkmark &  & \checkmark &  & $\times$ &  & .794 &  & .861 &  & .826 &  & .727 \\
(16) &  & DINOv2-B~\cite{oquab2023dinov2} &  & \checkmark &  & \checkmark &  & \checkmark &  & \checkmark &  & $\times$ &  & \textbf{.816} &  & \textbf{.862} &  & \textbf{.838} &  & \textbf{.744} \\
(17) &  & DINOv2-L~\cite{oquab2023dinov2} &  & $\times$ &  & \checkmark &  & \checkmark &  & \checkmark &  & $\times$ &  & .667 &  & .517 &  & .582 &  & .454 \\
(18) &  & DINOv2-L~\cite{oquab2023dinov2} &  & L &  & \checkmark &  & \checkmark &  & \checkmark &  & $\times$ &  & .803 &  & .843 &  & .823 &  & .725 \\
(19) &  & DINOv2-L~\cite{oquab2023dinov2} &  & \checkmark &  & \checkmark &  & \checkmark &  & \checkmark &  & $\times$ &  & .809 &  & .850 &  & .829 &  & .732 \\ 
(20) &  & DINOv3-B~\cite{dinov3} &  & $\times$ &  & \checkmark &  & \checkmark &  & \checkmark &  & $\times$ &  & .692 &  & .588 &  & .636 &  & .478 \\
(21) &  & DINOv3-B~\cite{dinov3} &  & L &  & \checkmark &  & \checkmark &  & \checkmark &  & $\times$ &  & .690 &  & .753 &  & .720 &  & .598 \\
(22) &  & DINOv3-B~\cite{dinov3} &  & \checkmark &  & \checkmark &  & \checkmark &  & \checkmark &  & $\times$ &  & .709 &  & .787 &  & .746 &  & .605 \\
(23) &  & DINOv3-L~\cite{dinov3} &  & $\times$ &  & \checkmark &  & \checkmark &  & \checkmark &  & $\times$ &  & .718 &  & .593 &  & .650 &  & .496 \\
(24) &  & DINOv3-L~\cite{dinov3} &  & \checkmark &  & \checkmark &  & \checkmark &  & \checkmark &  & $\times$ &  & .722 &  & .791 &  & .755 &  & .621 \\ \hline
(25) &  & DINOv2-B~\cite{oquab2023dinov2} &  & \checkmark &  & $\times$ &  & $\times$ &  & $\times$ &  & $\times$ &  & .644 &  & .605 & \textbf{} & .624 &  & .469 \\
(26) &  & DINOv2-B~\cite{oquab2023dinov2} &  & \checkmark &  & $\times$ &  & \checkmark &  & \checkmark &  & $\times$ &  & .699 &  & .666 & \textbf{} & .682 &  & .526 \\
(27) &  & DINOv2-B~\cite{oquab2023dinov2} &  & \checkmark &  & \checkmark &  & $\times$ &  & \checkmark &  & $\times$ &  & \textbf{.822} &  & .728 & \textbf{} & .772 &  & .664 \\
(28) &  & DINOv2-B~\cite{oquab2023dinov2} &  & \checkmark &  & \checkmark &  & \checkmark &  & $\times$ &  & $\times$ &  & .804 &  & .838 & \textbf{} & .821 &  & .719 \\
(29) &  & DINOv2-B~\cite{oquab2023dinov2} &  & \checkmark &  & \checkmark &  & \checkmark &  & \checkmark &  & $\times$ &  & .816 &  & \textbf{.862} & \textbf{} & \textbf{.838} &  & \textbf{.744} \\ \hline
(30) &  & DINOv2-B~\cite{oquab2023dinov2} &  & \checkmark &  & \checkmark &  & \checkmark &  & \checkmark &  & \checkmark &  & \textbf{.939} & \textbf{} & \textbf{.971} &  & \textbf{.954} &  & \textbf{.912} \\ \hline
\end{tabular}
\label{tab: sdgabl}
\end{table}

\begin{table}[]
\setlength{\tabcolsep}{1.35pt}
\caption{Ablation study on the threshold of QES.}
\begin{tabular}{ccccccccccccccccccccc}
\hline
QES Threshold  &  & 0.0  &  & 0.1  &  & 0.2  &  & 0.3  &  & 0.4  &  & 0.5  &  & 0.6  &  & 0.7  &  & 0.8  &  & 0.9  \\ \cline{1-1} \cline{3-3} \cline{5-5} \cline{7-7} \cline{9-9} \cline{11-11} \cline{13-13} \cline{15-15} \cline{17-17} \cline{19-19} \cline{21-21} 
kept sample ratio &  & 2.45 &  & 1.90 &  & 1.84 &  & 1.72 &  & 1.51 &  & 1.00 &  & 0.73 &  & 0.50 &  & 0.31 &  & 0.06 \\
IMD20 SDG IoU  &  & .744 &  & .825 &  & .842 &  & .865 &  & .889 &  & .912 &  & .924 &  & .957 &  & .967 &  & .968 \\ \hline
\end{tabular}
\label{tab: qesabl}
\end{table}

\begin{table}[t!]
\centering
\caption{Classification results for SDG, SPG and misaligned SPG. `Ensemble` denotes the ensemble of the models.}
\setlength{\tabcolsep}{3pt}
\begin{tabular}{ccccccccccccc}
\hline
\multirow{2}{*}{method} &  & \multicolumn{3}{c}{SDG} &  & \multicolumn{3}{c}{SPG} &  & \multicolumn{3}{c}{Misaligned} \\ \cline{3-5} \cline{7-9} \cline{11-13} 
 &  & P & R & F &  & P & R & F &  & P & R & F \\ \cline{1-1} \cline{3-5} \cline{7-9} \cline{11-13} 
DiNAT~\cite{dinat} &  & .992 & .992 & .992 &  & 1 & .996 & .998 &  & .867 & .929 & .897 \\
SwinTrans~\cite{liu2021swin} &  & .996 & .981 & .988 &  & 1 & .996 & .998 &  & .737 & 1 & .849 \\
ConvNeXt~\cite{liu2022convnet} &  & .996 & .992 & .994 &  & 1 & .996 & .998 &  & .875 & 1 & .933 \\
Ensemble &  & .996 & 1 & .998 &  & 1 & .996 & .998 &  & 1 & 1 & 1 \\ \hline
\end{tabular}
\label{tab: spgsdg_exp}
\end{table}

\begin{table}[t!]
\caption{Image manipulation localization performance on SACP~\cite{sacp} dataset. `Forgery Baseline` denotes using the tampCOCO~\cite{catnet} and CASIAv2~\cite{casia} datasets for model pretraining}
\setlength{\tabcolsep}{1.8pt}
\begin{tabular}{ccccccccc}
\hline
Num. &  & Methods &  & Pretraining data &  & IoU &  & F1 \\ \cline{1-1} \cline{3-3} \cline{5-5} \cline{7-7} \cline{9-9} 
(1) &  & DFCN~\cite{densefcn} &  & ImageNet~\cite{deng2009imagenet} &  & .466 &  & .607 \\
(2) &  & MVSS-Net~\cite{dong2022mvss} &  & ImageNet~\cite{deng2009imagenet} &  & .401 &  & .534 \\
(3) &  & PSCC-Net~\cite{liu2022pscc} &  & ImageNet~\cite{deng2009imagenet} &  & .482 &  & .620 \\
(4) &  & RRU-Net~\cite{rru} &  & ImageNet~\cite{deng2009imagenet} &  & .517 &  & .651 \\
(5) &  & CFL-Net~\cite{cflnet} &  & ImageNet~\cite{deng2009imagenet} &  & .433 &  & .571 \\
(6) &  & DTD~\cite{dtd} &  & ImageNet~\cite{deng2009imagenet} &  & .588 &  & .712 \\
(7) &  & TIFDM~\cite{tifdm} &  & ImageNet~\cite{deng2009imagenet} &  & .576 &  & .703 \\
(8) &  & Web-IML (Ours) &  & ImageNet~\cite{deng2009imagenet} &  & .773 &  & .862 \\
(9) &  & Web-IML (Ours) &  & ADE20k~\cite{ade20k} &  & .774 &  & .863 \\
(10) &  & Web-IML (Ours) &  & Forgery Baseline &  & .778 &  & .865 \\
(11) &  & Web-IML (Ours) &  & (10) + DEFACTO~\cite{defacto} &  & .767 &  & .856 \\
(12) &  & Web-IML (Ours) &  & (10) + PSCC-Synthetic~\cite{liu2022pscc} &  & .762 & \multicolumn{1}{l}{} & \multicolumn{1}{l}{.853} \\
(13) &  & Web-IML (Ours) &  & (10) + AIGC Synthetic~\cite{wang2025opensdi} &  & .756 & \multicolumn{1}{l}{} & \multicolumn{1}{l}{.851} \\
\hline
(14) &  & Web-IML (Ours) &  & (10) + MIMLv1~\cite{mimlv1} (\textit{Conference}) &  & .792 & \multicolumn{1}{l}{} & \multicolumn{1}{l}{.872} \\
\rowcolor{gray!15}(15) &  & Web-IML (Ours) &  & (10) + Web Supervision (Ours) &  & \textbf{.800} & \textbf{} & \textbf{.880} \\ \hline
\end{tabular}
\label{tab: sacp}
\end{table}

\subsection{Experiments on the Downstream Document IML Task}
This subsection evaluates the generalization of the Web-IML model and the universal applicability of our web-supervision method on an important downstream task: document IML.

We fine-tuned the Web-IML model, pretrained under various configurations, on the handcrafted real-world document IML benchmarks, SACP~\cite{sacp} and RTM~\cite{rtm}, adopting the same training and test configurations as previous studies~\cite{tifdm, rtm}. Across both benchmarks, our Web-IML model consistently and notably outperforms existing methods.
A key observation from Tables~\ref{tab: sacp} and~\ref{tab: rtm} is that settings 11, 12, and 13, which augmented the baseline (Setting 10) with additional pretraining data from DEFACTO~\cite{defacto}, PSCC~\cite{liu2022pscc}, and OpenSDID~\cite{wang2025opensdi}, respectively, exhibited degraded performance. This confirms that merely increasing the scale and diversity of pretraining data does not lead to improved real-world generalization.
In contrast, the model pretrained on our web-supervision data (Setting 15) demonstrates significant improvement. It achieved 2.2 points higher IoU on SACP and 4.7 points on RTM compared to the baseline (Setting 10). \textit{Furthermore, it also surpassed the model pretrained on MIMLv1 (Setting 14) by 0.8 points higher IoU on SACP and 1.5 points on RTM.}
These results demonstrate that the cost-effective data created by our web-supervision method possesses sufficient quality and relevance to enable models to learn robust semantic-agnostic forensic features, thereby generalizing effectively across diverse and unseen real-world IML tasks.
 
\begin{table}[]
\setlength{\tabcolsep}{2.4pt}
\caption{Image manipulation localization performance on RTM~\cite{rtm} dataset. `Forgery Baseline` denotes using the tampCOCO~\cite{catnet} and CASIAv2~\cite{casia} datasets for model pretraining.}
\begin{tabular}{ccccccccc}
\hline
Num. &  & Methods &  & Pretraining data &  & IoU &  & F1 \\ \cline{1-1} \cline{3-3} \cline{5-5} \cline{7-7} \cline{9-9} 
(1) &  & RRU-Net~\cite{rru} &  & ImageNet~\cite{deng2009imagenet} &  & .037 &  & .072 \\
(2) &  & PSCC-Net~\cite{liu2022pscc} &  & ImageNet~\cite{deng2009imagenet} &  & .033 &  & .064 \\
(3) &  & MVSS-Net~\cite{dong2022mvss} &  & ImageNet~\cite{deng2009imagenet} &  & .051 &  & .097 \\
(4) &  & CAT-Net~\cite{catnet} &  & ImageNet~\cite{deng2009imagenet} &  & .113 &  & .203 \\
(5) &  & DTD~\cite{dtd} &  & ImageNet~\cite{deng2009imagenet} &  & .065 &  & .122 \\
(6) &  & Mask2Former~\cite{cheng2021mask2former} &  & ADE20k~\cite{ade20k} &  & .124 &  & .172 \\
(7) &  & ASCFormer~\cite{rtm} &  & ADE20k~\cite{ade20k} &  & .198 &  & .329 \\
(8) &  & Web-IML (Ours) &  & ImageNet~\cite{deng2009imagenet} &  & .192 &  & .323 \\
(9) &  & Web-IML (Ours) &  & ADE20k~\cite{ade20k} &  & .207 &  & .343 \\
(10) &  & Web-IML (Ours) &  & Forgery Baseline &  & .194 &  & .325 \\
(11) &  & Web-IML (Ours) &  & (10) + DEFACTO~\cite{defacto} &  & .186 &  & .313 \\
(12) &  & Web-IML (Ours) &  & (10) + PSCC-Synthetic~\cite{liu2022pscc} &  & .189 &  & .318 \\
(13) &  & Web-IML (Ours) &  & (10) + AIGC Synthetic~\cite{wang2025opensdi} &  & .185 &  & .311 \\
(14) &  & Web-IML (Ours) &  & (10) + MIMLv1~\cite{mimlv1} &  & .226 &  & .369 \\
\rowcolor{gray!15}(15) &  & Web-IML (Ours) &  & (10) + Web Supervision (Ours) &  & \textbf{.241} & \textbf{} & \textbf{.389} \\ \hline
\end{tabular}
\vspace{-0.2cm}
\label{tab: rtm}
\end{table}

  \begin{figure}[t!]
    \centering
 	\includegraphics[width=1.0\linewidth]{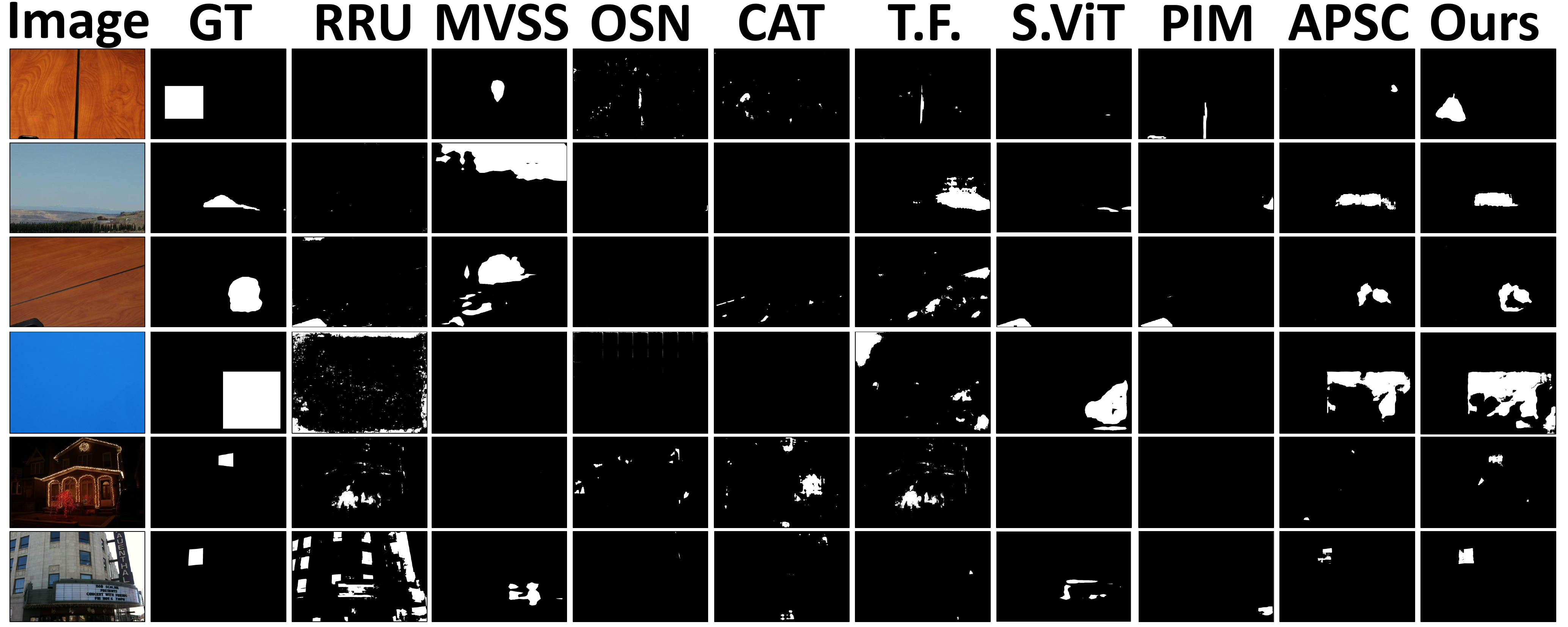}
 	\caption{Both Web-IML and previous methods sometimes struggle to produce precise mask when the forged region exhibits few visual artifacts. From left to right: input image, Ground-Truth (abbreviated as GT), RRU-Net (RRU), MVSS-Net (MVSS), IF-OSN (OSN), CAT-Net (CAT), TruFor (T.F.), SparseViT (S.ViT), APSC-Net (APSC) and ours.
 	\setlength{\abovecaptionskip}{-0.4cm}
 	} \label{fig:fa}
 \end{figure}

\section{Limitations}
Although our QES metric is highly effective in ensuring overall annotation quality, a small number of correct annotations may also be filtered out. In addition, our Web-IML is designed to maximize accuracy by leveraging web-scale supervision with a medium-sized model (140M). As a result, it is not highly computationally efficient (achieving 10 frames per second on a single 3090 GPU). Figure~\ref{fig:fa} presents some failure cases, showing that both Web-IML and previous methods may struggle to produce precise masks when the manipulated region exhibits very few visual artifacts.

\section{Conclusion}
In this paper, we addressed the critical challenge of data scarcity in image manipulation localization through a scalable web-supervision approach. At the core of our approach is CAAAv2, a novel category-aware and prior-denoising auto-annotation paradigm. This innovative design strategically separates manipulations into Shared Donor Group and Shared Probe Group, effectively converting the challenging forensic task into a prior-denoising task. This dramatically reduces task complexity and mitigates overfitting.
Leveraging this automatic annotation engine and our proposed QES metric for quality control, we constructed MIMLv2, a diverse, high-quality dataset of 246,212 manually forged images—over 120× larger than existing handcrafted datasets. 
We further enhanced this web-scale supervision with Object Jitter, a novel technique for generating challenging training examples from authentic images. To fully capitalize on this vast data, we designed Web-IML, a new model featuring multi-scale perception and self-rectification modules.
Extensive experiments demonstrate that our web-supervision method provides unprecedented performance gains across multiple models on a wide array of real-world forgery benchmarks. 
Additionally, Web-IML establishes a new state-of-the-art, significantly outperforming previous methods and demonstrating remarkable generalization. 
We believe that this work establishes a scalable framework for continuously expanding training data as new manually forged images emerge online. 
This research transforms image forensics, shifting reliance from limited, expensive handcrafted datasets to abundant, continuously growing web resources, thus paving the way for more robust image manipulation localization systems.

\bibliographystyle{IEEEtran}
\bibliography{IEEEabrv, mainbib}

\vfill

\end{document}